\pdfoutput=1

\documentclass[11pt]{article}

\usepackage[final]{acl}

\usepackage{times}
\usepackage{caption}
\usepackage{latexsym}
\usepackage[T1]{fontenc}
\usepackage{longtable}
\usepackage{tikz}
\usepackage{pgfplots}
\pgfplotsset{compat=1.17} 

\usepackage[utf8]{inputenc}
\usepackage{microtype}

\usepackage{svg}
\usepackage{amsmath}
\usepackage{algorithm}
\usepackage[noend]{algpseudocode}
\usepackage{graphicx}
\usepackage{svg}
\usepackage{amsmath}
\usepackage{float}
\usepackage{caption}
\usepackage{subcaption}
\usepackage{soul}
\usepackage{multirow}
\usepackage{multicol}
\usepackage{numprint}
\usepackage{longtable}
\usepackage{makecell}
\usepackage{booktabs, multirow} 
\usepackage{changepage, threeparttable} 

\definecolor{TreeGreen}{RGB}{46, 139, 87}
\definecolor{GoodGreen}{RGB}{3, 146, 94}
\definecolor{BadRed}{RGB}{250, 70, 70}

\graphicspath{ {./images/} }

\title{Semantically-Prompted Language Models Improve Visual Descriptions}

\author{Michael Ogezi, Bradley Hauer, and Grzegorz Kondrak \\
  Alberta Machine Intelligence Institute, \\
  Department of Computing Science, \\ 
  University of Alberta, Edmonton, Canada \\
  \href{mailto:ogezi@ualberta.ca}{\color{black}\texttt{{\{mikeogezi,bmhauer,gkondrak\}@ualberta.ca}}} \\
}

\begin{document}
\maketitle

\begin{abstract}
Language-vision models like CLIP have made significant strides in vision tasks, such as zero-shot image classification (ZSIC). 
However, generating specific and expressive visual descriptions remains challenging; descriptions produced by current methods are often ambiguous and lacking in granularity. 
To tackle these issues, we propose {V-GLOSS}: {V}isual {Gloss}es, a novel method built upon two key ideas. 
The first is Semantic Prompting, which conditions a language model on structured semantic knowledge. 
The second is a new contrastive algorithm that elicits fine-grained distinctions between similar concepts. 
With both ideas, we demonstrate that V-GLOSS improves visual descriptions and achieves strong results 
in the zero-shot setting
on general and fine-grained image-classification datasets, including ImageNet, STL-10, FGVC Aircraft, and Flowers 102.
Moreover, these descriptive capabilities contribute to enhancing image-generation performance.
Finally, we introduce a quality-tested silver dataset with descriptions generated with V-GLOSS for all ImageNet classes.
\end{abstract}

\begin{figure*}[t]
\centering
\begin{tabular}{@{}c@{\hspace{0.75cm}}c}
\includegraphics[width=0.45\textwidth]{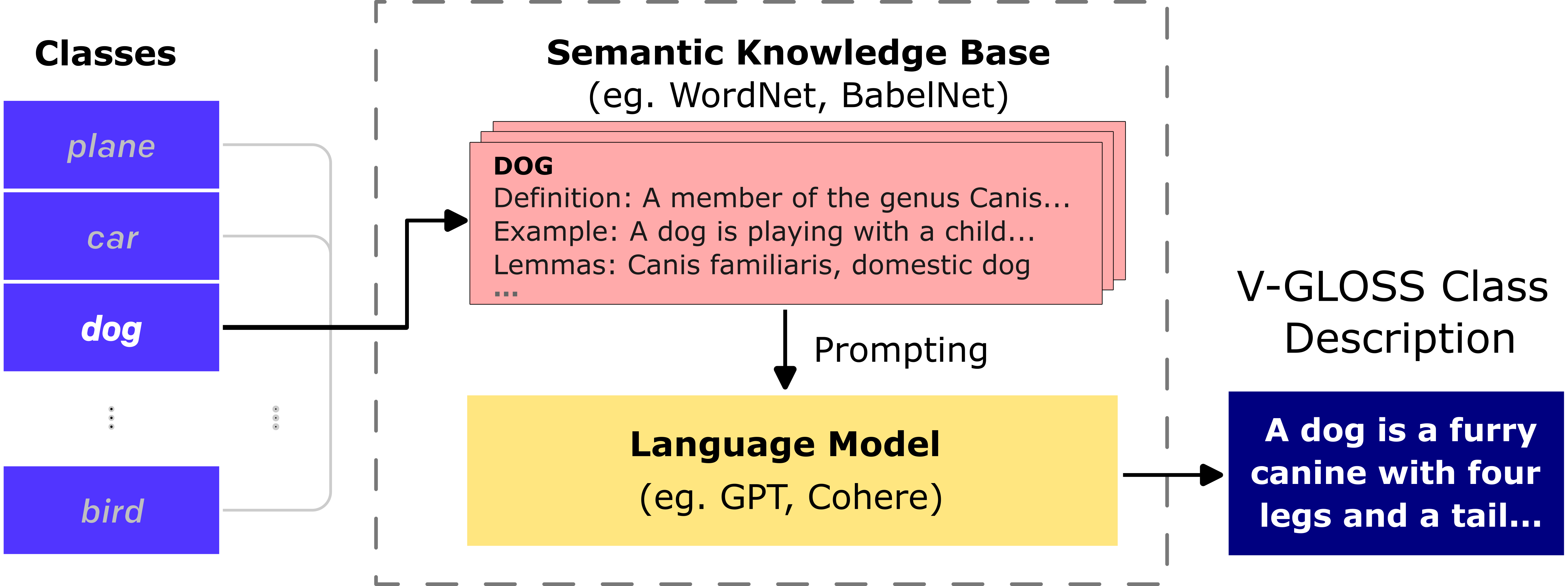} & \multirow[c]{2}{*}[1.3cm]{\includegraphics[width=0.5\textwidth]{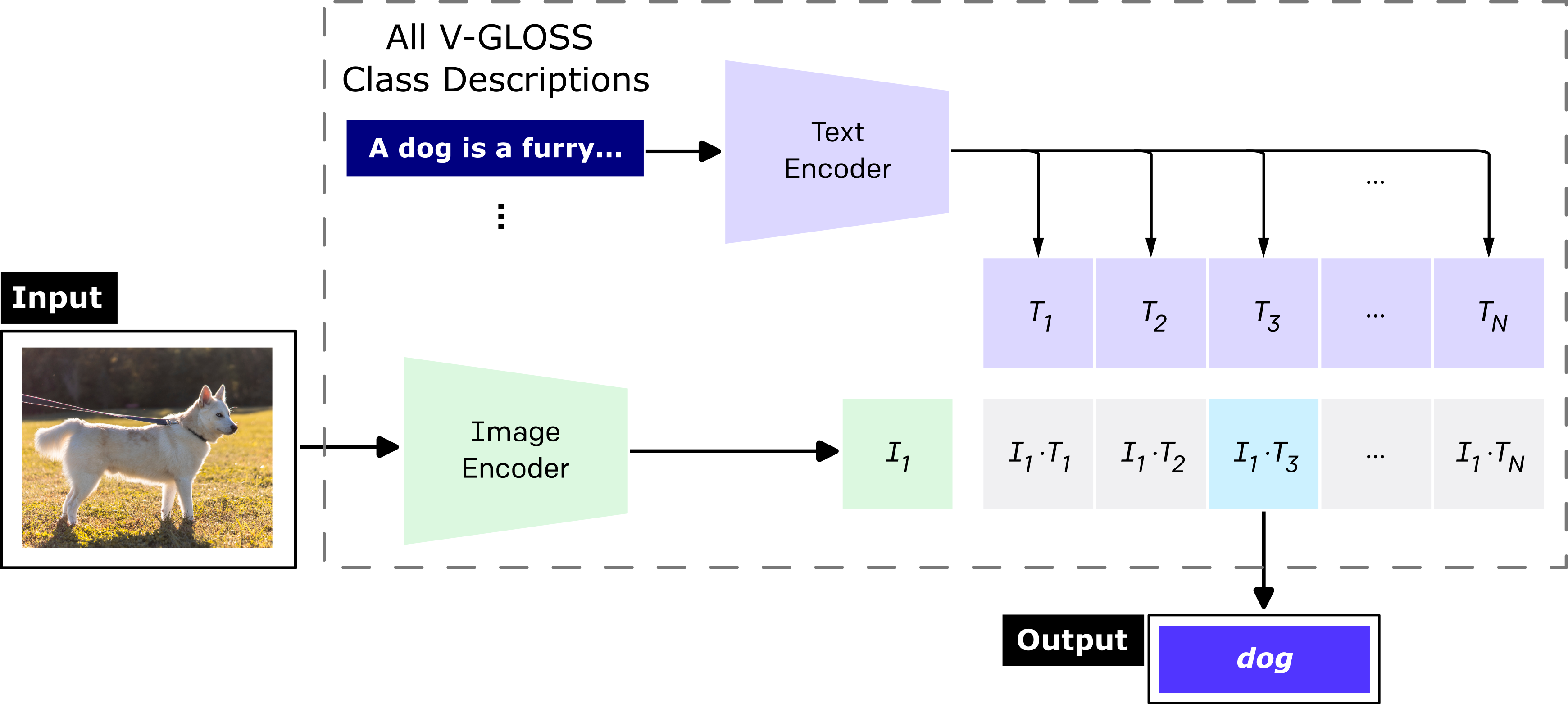}} \\
{\small (a) V-GLOSS producing a \textsc{dog} description} & \\
 & \\
\includegraphics[width=0.375\textwidth]{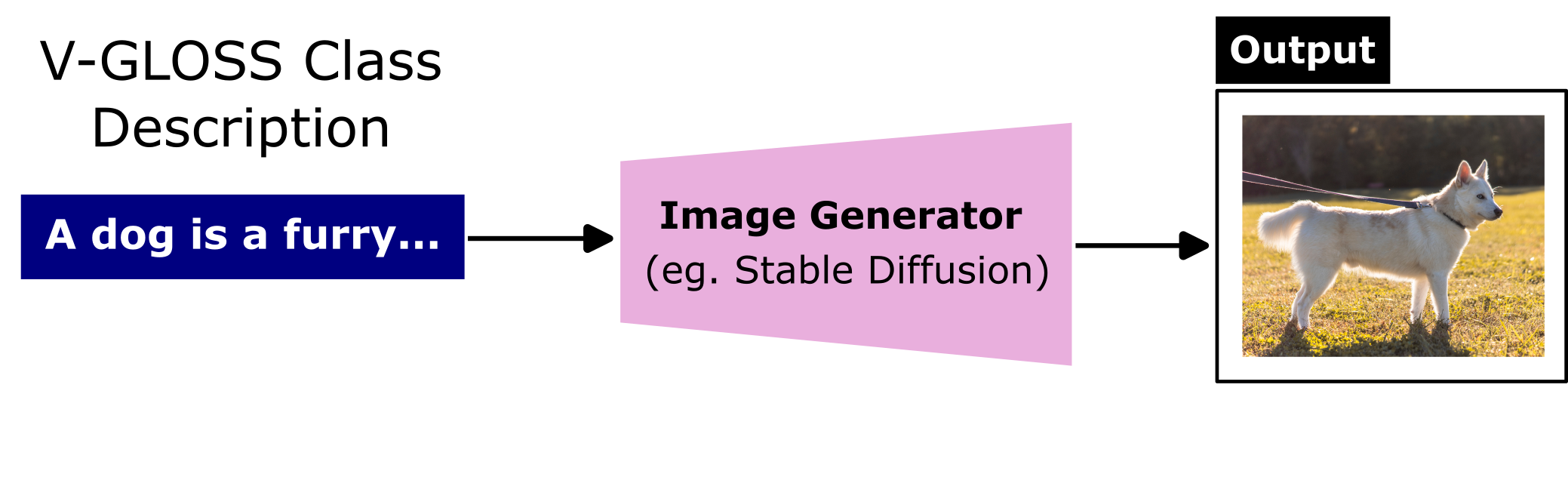} & \\
{\small (b) V-GLOSS for ZSCIG: generating a \textsc{dog} image} & {\small (c) V-GLOSS for ZSIC: classifying a test image} \\
\end{tabular}
\caption{
For the \textsc{dog} class, we depict (a) V-GLOSS's architecture (Section \ref{par:normal}), along with adaptations: (b) zero-shot image classification (ZSIC) (Section \ref{par:zsic}) and (c) zero-shot class-conditional image generation (ZSCIG) (Section \ref{par:zscig})}
\label{fig:classification}
\end{figure*}

\section{Introduction}
\label{sec:intro}

Language-vision models \cite{clip, align} have made significant progress 
in zero-shot vision tasks. 
However, in agreement with \citet{betker2023improving}, we hypothesize that their accuracy is limited by a lack of visual concept descriptions that are both expressive and specific,
that is, glosses that detail the unique visual characteristics of a concept. 
In this work, we investigate this hypothesis by creating and testing a new method for producing visual descriptions with pre-trained language models and semantic knowledge bases.

\begin{table}[t]
\centering
\small
\def\arraystretch{0.8}
\begin{tabular}{p{2.1cm} p{1.9cm} p{2.5cm}}
\toprule
{{Class / Concept}} & {{WordNet Gloss}} & {{V-GLOSS (Ours)}} \\
\midrule
\textsc{Corkscrew} \begin{center} \vspace*{0cm}\includegraphics[width=2cm] {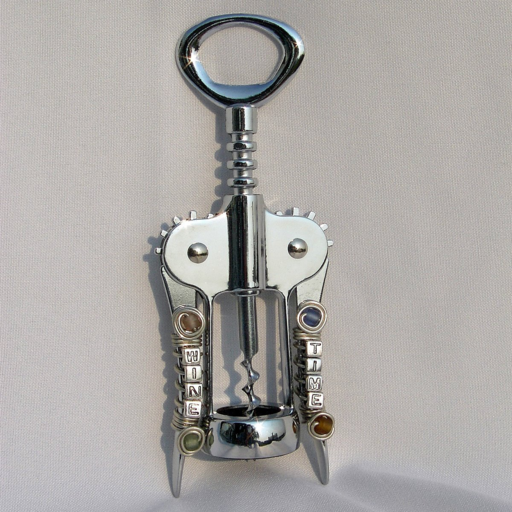}\vspace*{-0.4cm} \end{center} & \begin{flushleft} A bottle opener that pulls corks. \end{flushleft} & \begin{flushleft} A \textbf{tool} with a \textbf{spiral blade} that is used to remove corks from {bottles}. \end{flushleft} \\
\midrule
\textsc{Brambling} \begin{center} \vspace*{0cm}\includegraphics[width=2cm]{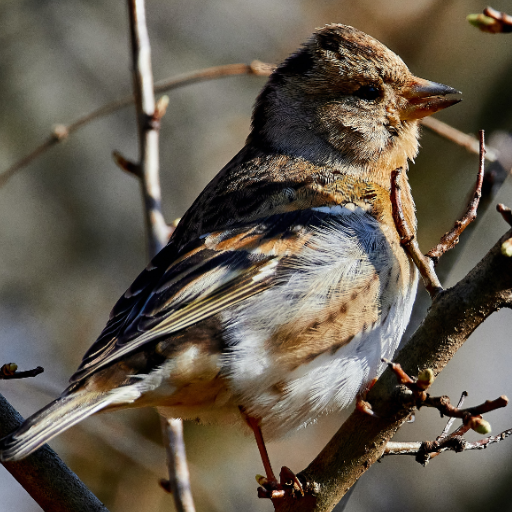}\vspace*{-0.4cm} \end{center} & \begin{flushleft} Eurasian finch. \end{flushleft} & \begin{flushleft} A \textbf{small brown} bird with a \textbf{black head} and a \textbf{white patch} on its chest. \end{flushleft} \\
\midrule
\textsc{Broccoli} \begin{center} \vspace*{0cm}\includegraphics[width=2cm]{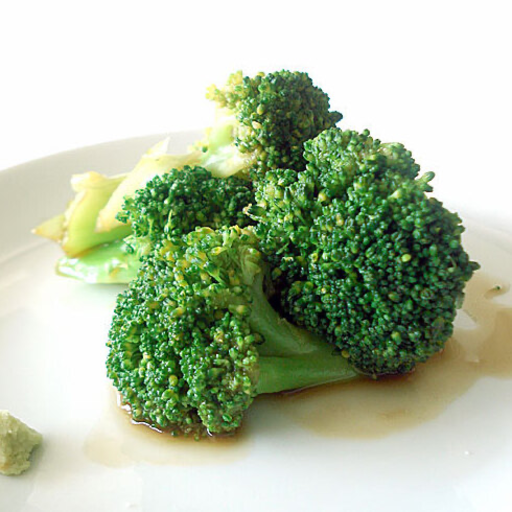}\vspace*{-0.4cm} \end{center} & \begin{flushleft} Branched green undeveloped flower heads. \end{flushleft} & \begin{flushleft} A \textbf{green vegetable} with a \textbf{thick stalk} and \textbf{florets} that grow in a \textbf{dense} head. \end{flushleft} \\
\bottomrule
\end{tabular}
\caption{A qualitative comparison between baseline glosses and V-GLOSS descriptions
for some ImageNet classes. 
Our method describes the \textit{visual characteristics} of a class, instead of what it \textit{does} or \textit{is}.
Many more examples are shown in Table \ref{tab:imagenet_comparison}.
}
\label{table:descript_comparison}
\end{table}

High-quality visual descriptions are crucial in tasks such as zero-shot image classification and text-based image retrieval.
Improved descriptions facilitate the creation of more useful representations. 
These are essential in producing robust and adaptable methods capable of understanding novel and specific visual attributes without re-training.

Existing approaches to generating visual descriptions, such as Template Ensembling \cite{clip} and CuPL \cite{cupl}, involve directly plugging class labels into fixed templates (e.g., {\em A photo of X)}, 
and prompting large language models such as InstructGPT \cite{instruct-gpt} 
to generate descriptions based on class labels
(e.g., {\em What does {X} look like?}), respectively.  
These methods suffer from two main issues: class granularity and label ambiguity. 
Class granularity refers to the difficulty in distinguishing between visually similar classes, 
such as \textsc{alligator} and \textsc{crocodile}. 
Label ambiguity 
is caused by using polysemous words as 
labels for distinct concepts. 
For example, 
\textsc{crane} can refer to either a 
bird or a construction machine. 
These issues limit the performance of existing models \cite{clip}. 

To address these challenges, we introduce 
\hbox{V-GLOSS}, 
a novel method  
that leverages language models (LMs) and semantic knowledge bases (SKBs) 
to generate 
improved visual descriptions
-- \textbf{V}isual \textbf{Gloss}es.
Table~\ref{table:descript_comparison} shows some examples.
By combining structured semantic information from SKBs such as WordNet \cite{wordnet98}, and BabelNet \cite{babelnet}, with a contrastive algorithm to finely distinguish similar classes, 
V-GLOSS is designed to mitigate the dual issues of granularity and ambiguity.

Our results demonstrate the effectiveness of V-GLOSS in improving the performance of ZSIC systems. 
We achieve strong improvements compared to prior work on benchmark datasets such as ImageNet \cite{imagenet} (+1.8\%), FGVC Aircraft \cite{fgvc_aircraft} (+2.6\%), and Flowers 102 \cite{flowers} (+1.6\%) in the zero-shot setting. 
Additionally, 
we introduce V-GLOSS Silver, 
a silver dataset constructed by V-GLOSS, 
which consists of a visual description for each ImageNet class. 
We show that V-GLOSS Silver is useful for zero-shot language-vision tasks 
such as ZSIC and ZSCIG,
comparing favorably to WordNet glosses. 


\section{Tasks}

Our main task is to generate a description for a given class or concept.
For example, if an image classification dataset has the class \textsc{dog}, 
we aim to produce a description such as \textit{``A dog is a furry, four-legged canine...''}
We consider such a description to be a specific kind of gloss. 

We use two downstream tasks to compare methods of generating class descriptions:
zero-shot image classification (ZSIC), and zero-shot class-conditional image generation (ZSCIG).
In ZSIC, the goal is to classify an image based on a set of classes, 
without having seen any labeled images belonging to those classes. 
The set of classes depends on the dataset.
For example, given an image 
depicting a dog, we aim to predict 
the class \textsc{dog}. 
%
In ZSCIG, the goal is to generate an image that corresponds to a specific class, 
again without having seen any labeled examples.
For example, given a class \textsc{dog}, we aim to generate an image of a dog.

In short, ZSIC is the task of classifying a given image,
while
ZSCIG is the task of generating an image given a class.
Both involve classes and images.
Visual descriptions of classes provide useful information which can facilitate both tasks 
by making it easier to either recognize or generate images of each class.
Therefore,
we aim to improve performance on both ZSIC and ZSCIG
by developing a novel method to improve the generation of such descriptions.

\section{Related Work}

\paragraph{Language Models}
The advent of transformer-based language models has revolutionized many natural language processing tasks \cite{gpt1, bert, gpt2, gpt3, gpt-neox, instruct-gpt}. 
As these models are scaled up by their number of parameters and quantity of training data, they exhibit emergent abilities such as few-shot and zero-shot learning \cite{emergent}. 

\paragraph{Language-Vision Models}
Significant strides have been made in the field of language-vision models such as CLIP \cite{clip} and ALIGN \cite{align}. 
These models apply contrastive pre-training approaches on large image-text datasets, leading to improved representation learning for both text and images and enhanced performance on several multi-modal tasks \cite{clipcap, clipfs}. 
Further advancements have been achieved by scaling up pre-training and incorporating auxiliary training objectives \cite{basic, coca}. 

\paragraph{Producing Descriptions \& Prompting}
The generation of descriptions and prompting has been explored in various studies. 
\citet{clip} introduced the template ensembling (TE) method, which uses a custom set of class labels and a fixed set of templates. 
Each label is inserted into these templates, and the completed templates for each class are aggregated into a single representation of the class. 
The CuPL method \cite{cupl} utilizes InstructGPT \cite{gpt3, instruct-gpt} to generate descriptions for ImageNet classes. 
Both TE and CuPL can be used for zero-shot image classification. \citet{optimizing} fine-tuned GPT models \cite{gpt1, gpt2} to rephrase image-generation prompts, resulting in improved images. 
\cite{coop} learned soft prompts that improve performance, but are intractable to humans. 
In this work, we prompt language models with semantic knowledge to generate visual descriptions.

\section{Method}

We begin by describing how we map classes 
to concepts in a semantic knowledge base (SKB), 
to leverage the concept-specific information the SKB contains. 
We then introduce our novel method V-GLOSS,
which has two variants,
\textit{normal} and \textit{contrastive}.
We conclude by describing the construction of V-GLOSS Silver,
a set of class descriptions produced using V-GLOSS.

\begin{figure}[H]
  \centering

  \vspace{0.2cm}
  \begin{subfigure}[][][t]{0.5\textwidth}
         \centering
         {\includegraphics[width=1.0\textwidth]{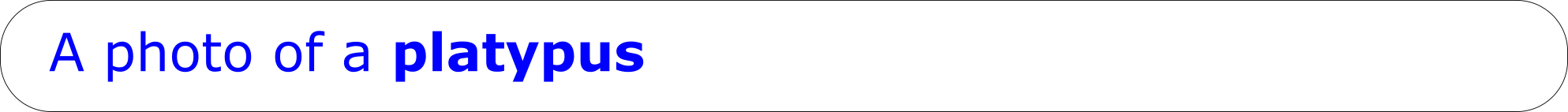}}
         \caption{
         CLIP \cite{clip}
         }
         \label{fig:baseline}
  \end{subfigure}

  \vspace{0.2cm}  
  \begin{subfigure}[][][t]{0.5\textwidth}
         \centering
         {\includegraphics[width=1.0\textwidth]{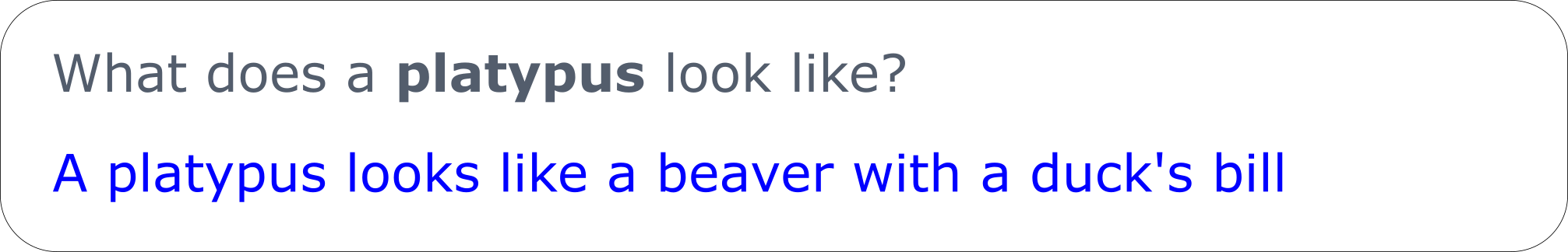}}
         \caption{
         CuPL \cite{cupl}
         }
         \label{fig:cupl}
  \end{subfigure}

  \vspace{0.2cm}
  \begin{subfigure}[][][t]{0.5\textwidth}
         \centering
         {\includegraphics[width=1.0\textwidth]{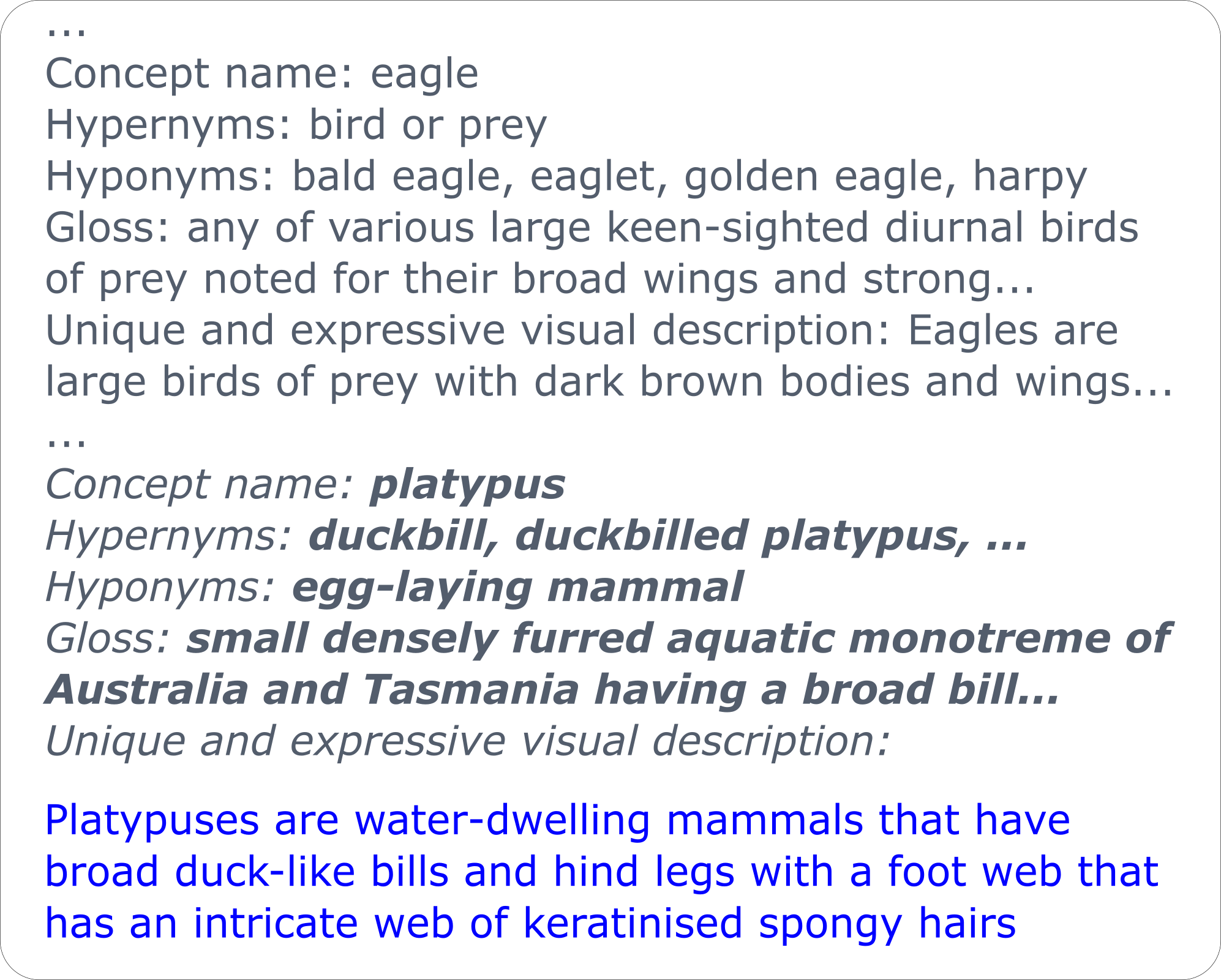}}
         \caption{
         V-GLOSS (Ours)
         }
         \label{fig:ours}
  \end{subfigure}
  \caption{
  Class descriptions for \textsc{Platypus} 
  produced by one template-based method (a) and two that use LMs (b and c).
  {\color{gray} Input} prompts, {\color{blue} output} descriptions, and \textbf{plugged values} are shown.
  }
  \label{fig:method}
\end{figure}

\subsection{Mapping Classes to Synsets}
\label{sec:WNmap}

The ImageNet classes are already mapped to WordNet synsets
by the dataset's creators.
For the other datasets, we employ a heuristic that starts 
by mapping each class to the most frequent sense of the class label, 
as determined by WordNet\footnote{\url{https://www.nltk.org/}}. 
For CIFAR-10 and STL-10, this heuristic is sufficient.
For CIFAR-100, we 
manually re-map 18 classes.
For instance, we needed to re-map \textsc{ray} from \textit{light} to \textit{sea creature}, as the \textit{light} sense is most frequent,
but the \textsc{ray} in the dataset refers to the sea creature. We show mis-mapped CIFAR-100 classes in Table \ref{tab:classes_senses} of the appendix.
We fall back to our manually-produced definitions if no suitable synset is found in WordNet or BabelNet.
This happens 8 times for all 1,322 classes across all datasets, with all occurrences coming from FGVC Aircraft.


\subsection{V-GLOSS}
We discuss the two variants of V-GLOSS below, \textit{normal} and \textit{contrastive}. 
In both, for each class, we produce multiple descriptions resulting in an ensemble. Ultimately, to achieve our best results with V-GLOSS (\textit{Normal + Contrastive}) in Table \ref{table:main}, we combine both normal and contrastive, by concatenating the descriptions from each sub-method.
Unless otherwise stated, \emph{V-GLOSS} refers to this hybrid method.

\subsubsection{\textit{Normal} V-GLOSS}
\label{par:normal}

We generate normal descriptions via in-context learning with an LM,
beginning by providing the LM with a description of the task to be performed,
followed by multiple input-output examples. 
The examples are fixed, involving the concepts 
\textsc{eagle}, 
\textsc{\color{black}{bat}} (animal), 
\textsc{\color{black}{bat}} (baseball), and 
\textsc{television}. 
We selected these to expose the model to ambiguous class labels (\textit{bat}), 
a natural object (\textit{eagle}), 
and an artificial object (\textit{television}).
For each class,
we obtain the hypernyms, hyponyms, usage examples, synonyms, and gloss
of the sense to which the class is mapped,
and provide this to the LM.
Figure~\ref{fig:ours} shows a session with the LM, beginning with the
example of \textit{eagle}, with output generated for the class \textit{platypus}.
Table \ref{table:descript_comparison} compares our descriptions to baseline glosses. 

\subsubsection{\textit{Contrastive} V-GLOSS}
\label{par:contrastive}


\begin{figure}[t]
  \centering
  {\includegraphics[width=0.475\textwidth]{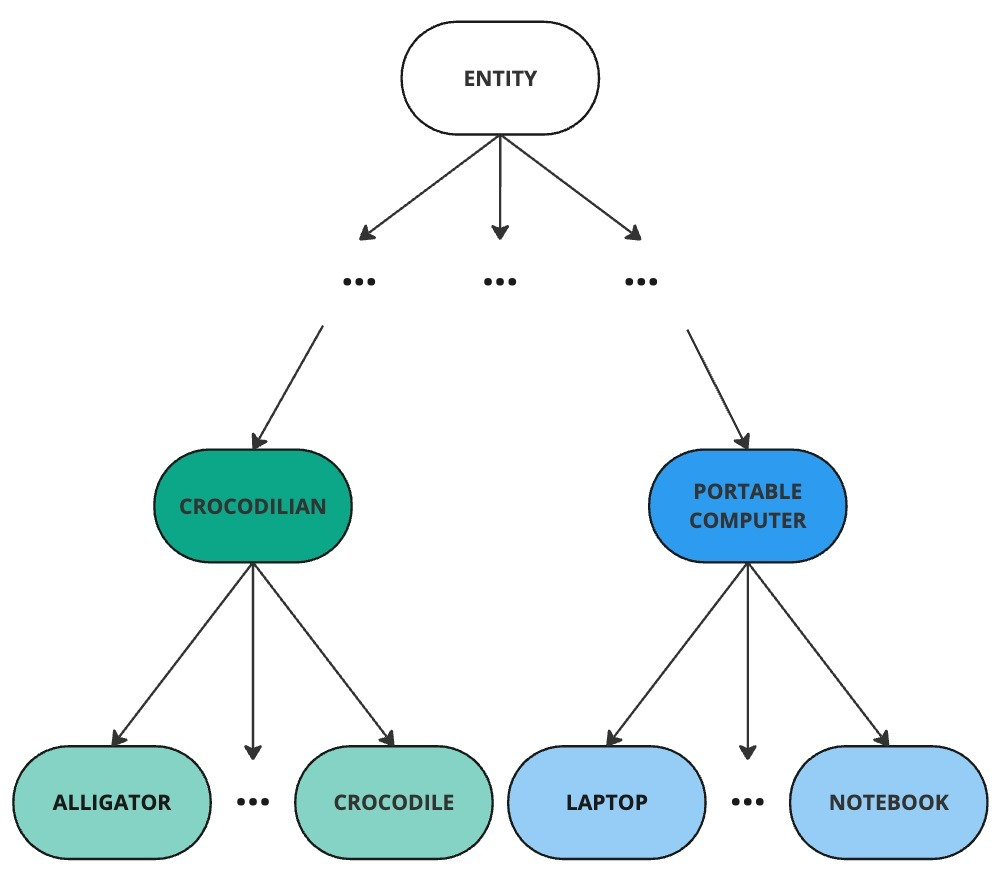}} 
  \caption{A sample of an SKB hypernym hierarchy.
  For \textit{contrastive} prompting, we only distinguish classes that are semantically  
  similar to the target class, like
  {\color{TreeGreen}\textsc{alligator}} to {\color{TreeGreen}\textsc{crocodile}}.}
  \label{fig:contrastive_tree}
\end{figure}

During development, 
we observed that many errors were caused by false positives involving visually similar classes. 
For example, 
the classes \textsc{crocodile} for \textsc{alligator}
refer to similar-looking animals,
and are often confused with one another.
Moreover, ImageNet contains 120 distinct dog species, and the fine-grained datasets contain only airplanes (FGVC Aircraft) or flowers (Flowers 102).
The contrastive variant of V-GLOSS is designed to address these issues by
using semantic similarity between classes as a heuristic to estimate visual similarity. 
For each class, we search for other classes that are semantically similar,
and if any are found, we add a negative instruction to the LM prompt,
e.g. we generate a description for an \textsc{alligator} \emph{but not} a \textsc{crocodile},
using the same prompt structure as for normal V-GLOSS.

We create a similarity matrix $M$ as follows:


\begin{equation}
    \label{eqn:matrix}
    M_{i,j} = Sim(S[i], S[j])
\end{equation}

$Sim(s_1, s_2)$ 
is the Wu-Palmer path-similarity function \cite{wu_palmer} 
comparing synsets $ s_1 $ and $ s_2 $; 
this similarity function uses the path between two concepts
in the WordNet tree (Figure \ref{fig:contrastive_tree}) to measure
semantic relatedness.
$S$ is the set of all classes in a dataset, $\mathcal{D}$, 
and $i$ and  $j$ are indices ranging from $1$ to $|S|$.
Concisely, Equation \ref{eqn:matrix} defines 
a similarity matrix containing similarity scores between all classes in a dataset. 
$ M $ is one of the inputs to our contrastive V-GLOSS variant, 
shown in Algorithm \ref{alg:contrastive}.


\begin{table}[t]
\centering
\small
\def\arraystretch{1.0}
\begin{tabular}{p{2.1cm} p{1.9cm} p{2.5cm}}
\toprule
\multirow{1}{*}{{Class / Concept}} & \textit{Normal} & \textit{Contrastive} \\
\midrule
\textsc{Alligator} \begin{center} \vspace*{0cm}\includegraphics[width=2cm]{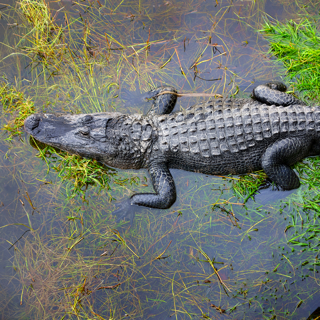}\vspace*{-0.4cm} \end{center} & \begin{flushleft} A large reptile with a long snout, a broad head, and a long tail. \vspace*{-0.4cm} \end{flushleft} & \begin{flushleft} A large, \textbf{dark-colored} reptile with a \textbf{rounded snout}, found in \textbf{freshwater}. \vspace*{-0.4cm} \end{flushleft} \\
\midrule
\textsc{Crocodile} \begin{center} \vspace*{0cm}\includegraphics[width=2cm]{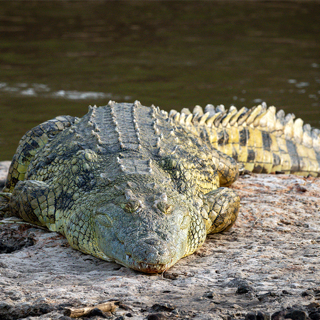}\vspace*{-0.45cm} \end{center} & \begin{flushleft} A reptile with a broad, flat snout, a long tail, and a long, pointed snout. \end{flushleft} & \begin{flushleft} A \textbf{grayish-green} reptile with a \textbf{v-shaped snout}, found in \textbf{brackish} or \textbf{saltwater}. \end{flushleft} \\
\bottomrule
\end{tabular}
\caption{
Two similar classes 
with \textbf{key differences}
between their \textit{normal} and \textit{contrastive} descriptions.
}
\label{table:constrastive_descriptions}
\end{table}

\begin{figure}[t]
\centering
\begin{tikzpicture}
\begin{axis}[
    width=\linewidth, 
    height=0.75\linewidth, 
    legend style={at={(0.5,-0.15)},
        anchor=north,legend columns=-1},
    symbolic x coords={5,10,15,20,25,30},
    xtick=data,
    xtick pos=bottom,
    ytick pos=left,
    axis lines=left,
    ymin=76,ymax=79, 
    ylabel={Accuracy (\%)},
    xlabel={$N$ (\emph{Contrastive} Prompts)},
    enlarge x limits=0.15
]
\addplot+[mark=*, gray, mark options={fill=black}] coordinates {(5,77.6) (10,78.0) (15,78.3) (20,78.5) (25,77.8) (30,76.5)};
\end{axis}
\end{tikzpicture}
\caption{V-GLOSS Accuracy vs $N$, with the number of \emph{normal} fixed at 50.}
\label{fig:accuracy_vs_N}
\end{figure}
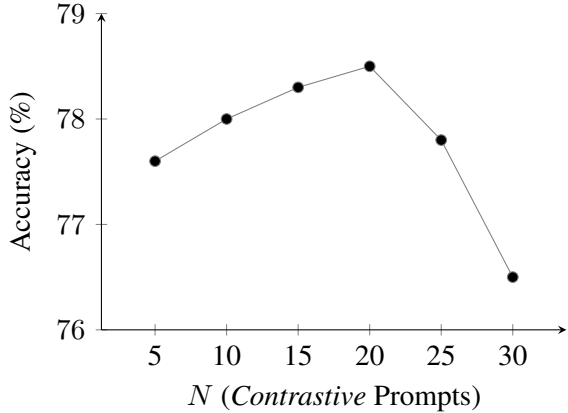

In Algorithm \ref{alg:contrastive}, $ \lambda $ is a threshold for minimum similarity. 
We only generate contrastive descriptions when classes have a similarity that exceeds or is equal to $ \lambda $.
$N$ indicates the maximum number of classes to generate contrastive descriptions for. To select $N$, we run a hyperparameter search (shown in Figure \ref{fig:accuracy_vs_N}).
$ k $ is the number of distinct descriptions to generate for a class pair.
$ LM_c $ takes in the $ target $ class, a neighbor class, and $ k $, then prompts the LM to generate $ k $ descriptions that distinguish the $ target $ and neighbor classes.
In summary, for each class, Algorithm \ref{alg:contrastive} identifies the classes most similar to it, excluding itself, and generates descriptions that distinguish them.
Table \ref{table:constrastive_descriptions} compares the normal and contrastive descriptions
for \textsc{alligator} and \textsc{crocodile};
note that distinguishing features of the two classes are included in the LM's output.
Table \ref{table:class_similarity} shows examples of classes with high false positive rates,
and the classes they are contrasted with.

\begin{algorithm}
    \caption{Generate \emph{Contrastive} Descriptions: We generate contrastive descriptions to help distinguish the most similar classes.}
    \label{alg:contrastive}
    \begin{algorithmic}[1]
        \Require $M$: Equation \ref{eqn:matrix} result
        \Require $\lambda$, $N$, $k$: Hyperparameters
        \Require $ S $: All classes in dataset, $ \mathcal{D} $
        \Require $ LM_c $: LM prompted contrastively 
        \State $G \gets$ empty $|S|$-list for class descriptions
        \For{$i \gets 0$ to $|S| - 1$}
        \State $target \gets S[i]$
        \State $S^* \gets $ top $N$ classes $ : \lambda \leq M_{i,*} \leq 1 $
            \For{$s^*$ in $S^*$}
                \State $ samples \gets LM_c(target, s^*, k) $
                \State $G[i].insert(samples)$
            \EndFor
        \EndFor
        \State \Return $ G $
    \end{algorithmic}
\end{algorithm}

\begin{table}[t]
\setlength{\parskip}{0pt}
\setlength{\parsep}{0pt}
\centering
\def\arraystretch{1}
\small
\begin{tabular}{p{1.5cm}p{2.75cm}p{2.15cm}}
\toprule
{Class} & {False Positives} & Contrastives \\
\midrule
\begin{flushleft} \vspace*{-0.3cm} \textsc{African Elephant} \vspace*{-0.6cm} \end{flushleft} & \begin{flushleft} \vspace*{-0.3cm} \textsc{Tusker} (44), \textsc{Asian Elephant} (6) \vspace*{-0.6cm} \end{flushleft} & \begin{flushleft} \vspace*{-0.3cm} {\color{GoodGreen} \textsc{Tusker}}, {\color{GoodGreen} \textsc{Asian Elephant}} \vspace*{-0.6cm} \end{flushleft} \\
\midrule
\begin{flushleft} \vspace*{-0.3cm} \textsc{Notebook} \vspace*{-0.3cm} \end{flushleft} & \begin{flushleft} \vspace*{-0.3cm} \textsc{Laptop} (22), \textsc{Desktop} (10), \textsc{Space bar} (2) \vspace*{-0.3cm} \end{flushleft} & \begin{flushleft} \vspace*{-0.3cm} {\color{GoodGreen} \textsc{Laptop}}, {\color{GoodGreen} \textsc{Desktop}}, {\color{BadRed} \textsc{Space bar}} \vspace*{-0.3cm} \end{flushleft} \\
\bottomrule
\end{tabular}
\caption{
False positives and their counts vs.\@  classes selected by the contrastive algorithm (see Equation \ref{eqn:matrix} and Algorithm \ref{alg:contrastive}).
{\color{GoodGreen} Hits} and {\color{BadRed} misses} are shown.
}
\label{table:class_similarity}
\end{table}

\section{Evaluation} 

In this section, we present our evaluation of V-GLOSS,
alongside comparable methods.
We describe our datasets, evaluation metrics, baselines, previous methods, and experiments. 
To ensure robustness, we report the mean over five random seeds in Tables \ref{table:visual_descriptions} and \ref{table:main}.

\subsection{Datasets}
\label{sec:eval_datasets}

We evaluate our method on the test splits of six widely used benchmark datasets, taking note to consider both general and fine-grained datasets. 
%

\paragraph{ImageNet} \cite{imagenet} consists of 50,000 images equally distributed across 1,000 classes, and serves as our {primary} benchmark. 
        
\paragraph{CIFAR-10} and \textbf{CIFAR-100} \cite{cifar} both comprise 10,000 test samples across 10 and 100 classes, respectively. 
        
\paragraph{STL-10} \cite{stl10} comprises 100,000 test samples designed for unsupervised learning.
        
\paragraph{FGVC Aircraft} \cite{fgvc_aircraft} contains 3,333 images across 100 aircraft model variants, with $\sim$33 images per variant.
        
\paragraph{Flowers 102} \cite{flowers} features 102 flower categories common in the UK, with 40 to 258 images per category.

\paragraph{}
For CIFAR-10, CIFAR-100, STL-10, FGVC Aircraft, and Flowers 102, which are not pre-mapped to WordNet, 
we employ the two-step process detailed in Section \ref{sec:WNmap} to map each class to a synset.

Experiment 1 (Section \ref{subsection:ex_1}) involves ImageNet alone and covers both the ZSCIG and ZSIC tasks.
In contrast, Experiment 2 (Section \ref{subsection:ex_2}), 
our main experiment, 
tests the impact of various class description methods on the ZSIC task 
and uses all datasets. 
In Experiment 2, we allow methods to use ensembles of descriptions of each class,
while in Experiment 1, we experiment with only a single description.




We selected these datasets to evaluate the following properties of V-GLOSS:
\begin{enumerate}
    \item 
        \textbf{Performance on common benchmark datasets with varying numbers of classes.} 
        Each dataset has its own set of classes,
        ranging from ImageNet with 1,000 classes, 
        to CIFAR-100 with 100 classes, 
        to CIFAR-10 and STL-10, each with 10 classes.
    \item 
        \textbf{Proficiency in fine-grained conceptual distinctions.}
        Although some datasets (ImageNet and CIFAR) cover diverse domains, we importantly consider fine-grained datasets like FGVC Aircraft and Flowers 102. This enables testing our method's ability to distinguish very similar classes (e.g., distinguishing between closely related species or types).
\end{enumerate}

\subsection{Evaluation Metrics}

\paragraph{Top-1 Accuracy}
In ZSIC, this metric is the frequency with which the model's top prediction for an image matches the gold label.

\paragraph{Fréchet Inception Distance (FID)}
For ZSCIG, FID \cite{fid} quantifies the divergence between ground truth and generated images, 
with lower scores signifying a better ability to produce images similar to the ground truth.

\paragraph{Inception Score}
Also for ZSCIG, the inception score \cite{inception_s} uses an Inception model's \cite{inception} output probability distribution to assess the diversity and realism of generated images, with higher scores indicating more diverse and convincing images.
Unlike the above metrics, this does not require ground-truth images for comparison.

\subsection{Baseline \& Previous Methods}
\label{subsec:baselines_plus}

In this section, we describe the methods to which we compare V-GLOSS.
For methods that produce ensembles of class descriptions
(i.e. multiple descriptions per class), 
a single representation of the class is obtained by averaging
individual representations for each description. 

First, the \textbf{1-Template baseline} inserts a class label into a \textit{single} specific template. 
For example, given the class \textsc{dog}, the baseline produces \textit{``A photo of a dog.''}

The next approach we consider is \textbf{Template Ensembling} \cite{clip},
which generates an ensemble of descriptions for a class by
inserting the class label into each of a set of 80 templates. 
For example, some descriptions for \textsc{dog} are: 
\textit{``A photo of a dog.''}, 
\textit{``A blurry photo of a dog.''}, 
and \textit{``An origami dog.''} 
This method uses a modified list of class labels\footnote{\url{https://github.com/anishathalye/imagenet-simple-labels}} 
designed to reduce ambiguity.

\textbf{CuPL} \cite{cupl}
also generates an ensemble of descriptions for each class.
The descriptions are generated by prompting a LLM, 
InstructGPT \cite{instruct-gpt},
with questions such as: 
\textit{``What does a dog look like?''} 
and \textit{``Describe an image of a dog from the internet.''} 
CuPL uses the same class labels as Template Ensembling.


\subsection{Experiment 1: V-GLOSS Silver}
\label{subsection:ex_1} 

This experiment evaluates V-GLOSS's ability to generate a \textit{single} description for each class,
without relying on ensembling.
We then evaluate the V-GLOSS description of each class against its WordNet gloss. 


To construct this set of class descriptions, 
which we view as a silver dataset of such descriptions,
we generate a single, normal description for each ImageNet class via greedy decoding. 
We generate only normal descriptions 
because they outperform contrastive ones 
when only a single description is used.
We call the resulting dataset \emph{V-GLOSS Silver}.

We extrinsically evaluate V-GLOSS Silver by using it for the ZSIC and ZSCIG tasks, 
and comparing the results to those achieved using
the 1-Template baseline, and WordNet glosses. 
We do not compare V-GLOSS Silver to CuPL or other previous methods 
which may produce more than one gloss for each class.

\subsubsection{Technical Details}

\paragraph{ZSIC} 
\label{par:zsic}
We employ CLIP \cite{clip}, which comprises an image encoder and a text encoder, as the ZSIC backbone model.
Our procedure consists of three steps:
First, we use the CLIP text encoder to create an aggregate representation for each class based on its 
description(s). 
Then, at test time, we employ the CLIP image encoder to generate a representation of the input image. 
Finally, we predict the class which maximizes the cosine similarity 
between the representation of its description(s),
and the image representation (see Figure \ref{fig:classification}c).
We evaluate the predictions using top-1 accuracy.

\paragraph{ZSCIG} 
\label{par:zscig}
For ZSCIG (see Figure \ref{fig:classification}b), 
we condition Stable Diffusion \cite{stable} on each class description
before generating an image.
We use a guidance scale of $7.5$ and run $50$ diffusion steps. 
We evaluate the generated images using Inception and FID scores. 

\subsubsection{Results}

\begin{table}[t]
\centering
\small
\def\arraystretch{1}
\begin{tabular}{@{}llll@{}}
\toprule
& ZSIC & \multicolumn{2}{c}{ZSCIG} \\
\cmidrule(l){2-2} \cmidrule(l){3-4}
& Accuracy ↑ & Inception ↑ & FID ↓ \\
\midrule
Baseline (1-Template) & 71.0 & 99.7 & 25.7 \\
\midrule
WordNet Glosses & 44.7 & 58.5 & 30.0 \\
\midrule
V-GLOSS Silver & \textbf{72.3} & \textbf{109.6} & \textbf{20.0} \\
\bottomrule
\end{tabular}
\caption{
    Extrinsic evaluation on the tasks of ZSIC and ZSCIG. 
    \textbf{↓} means that lower is better.
}
\label{table:visual_descriptions}
\end{table}

The results of Experiment 1 are shown in Table \ref{table:visual_descriptions}. 
Based on our extrinsic evaluation on the ZSIC and ZSCIG tasks, 
\textit{V-GLOSS Silver} descriptions yield better performance
compared to baseline and WordNet Glosses. 
On ZSIC, we improve accuracy by 1.3\%; 
on ZSCIG, we improve Inception and FID scores by 9.9 and 5.7, respectively.
This demonstrates the effectiveness and utility of V-GLOSS:
our visual descriptions yield better results
on ZSIC and ZSCIG.

\begin{table*}[t]
\centering
\small
\def\arraystretch{1}
\begin{tabular}{@{}lllllllll@{}}
\toprule
\multirow{2.5}{*}{Method} & \multirow{2.5}{*}{Model} & \multicolumn{6}{c}{Accuracy (\%) on Datasets} & \multirow{2.5}{*}{\thead{\# LM \\ Parameters}} \\
\cmidrule(l){3-8}
& & {ImageNet} & \thead{CIFAR \\ 100} & \thead{CIFAR \\ 10} & \thead{STL \\ 10} & \thead{FGVC \\ Aircraft} & \thead{Flowers \\ 102} \\
\midrule
\multirow{2}{*}{1-Template Baseline} 
         & {ViT} & 72.4 & 77.3 & 95.2 & 99.5 & 31.7 & 77.6 & \multirow{2}{*}{0} \\
         & {RN50} & 68.7 & 57.7 & 81.0 & 98.4 & 27.4 & 71.6 \\
\midrule
\multirow{2}{*}{Template Ensembling}
         & {ViT} & 76.2 & 77.9 & 96.2 & 99.4 & 32.9 & 78.5 & \multirow{2}{*}{0} \\
         & {RN50} & 73.2 & 61.3 & 86.8 & 98.3 & 29.7 & 74.3 \\
\midrule
\multirow{1}{*}{CuPL}
         & {ViT} & 76.7 & \textbf{78.6} & 95.8 & - & 36.1 & 79.7 & \multirow{1}{*}{175B} \\
\midrule
\multirow{2}{*}{V-GLOSS (\textit{Normal}-Only)}
         & {ViT} & 77.3 & 77.5 & 95.6 & 99.4 & 33.2 & 79.2 & \multirow{2}{*}{6.1B} \\
         & {RN50} & 73.3 & 63.5 & 86.8 & 98.3 & 30.8 & 75.1 \\
\midrule
\multirow{2}{*}{V-GLOSS (\textit{Normal + Contrastive})} 
         & {ViT} & \textbf{78.5} & {78.2} & \textbf{97.0} & \textbf{99.6} & \textbf{38.7} & \textbf{81.3} & \multirow{2}{*}{6.1B} \\
         & {RN50} & 74.5 & 64.6 & 87.8 & 98.8 & 35.2 & 77.3 \\
\bottomrule
\end{tabular}
\caption{
    Top-1 accuracy on ZSIC.  
    \underline{ViT}-L14-336 and \underline{RN50}x64 are Transformer- and ResNet-based CLIP variants.
    See Table \ref{table:main_full} for more model variants.
}
\label{table:main}
\end{table*}

\subsubsection{Analysis}

V-GLOSS Silver descriptions are considerably more detailed, more expressive, and better grounded than their WordNet counterparts (see Figure \ref{table:descript_comparison}). 
Specifically, we observe that V-GLOSS descriptions make greater use of descriptive words and phrases, e.g. {\em spiral, brown, green, thick, small}, etc. 

\subsection{Experiment 2: ZSIC}
\label{subsection:ex_2}

Our second experiment assesses the effectiveness of V-GLOSS descriptions in facilitating ZSIC.
The details for the ZSIC pipeline are largely similar to those described in Experiment 1 (Section \ref{subsection:ex_1}), 
except that we generate an ensemble of descriptions per class, as opposed to only one description.
%
We also experiment with two image encoder variants: ViT \cite{vit} and RN50 \cite{resnet}.
For all baselines and methods (Section \ref{subsec:baselines_plus}, Section \ref{par:normal}), 
we follow the same evaluation procedure after generating class descriptions. 

\subsubsection{Technical Details}
We generate class descriptions using the 6.1B-parameter Cohere LM\footnote{\url{https://docs.cohere.com/docs/models}}.
We choose Cohere over alternatives due to its extensive free plan, reducing the cost of our experiments. 
Cohere has comparable performance to the similarly-sized InstructGPT \cite{gpt3, instruct-gpt} variant, 
as demonstrated by \citet{helm} across various benchmarks. 
Therefore, we do not gain any advantage by using Cohere instead of InstructGPT.

When generating class descriptions with normal V-GLOSS, 
we use a temperature of $2.5$ to produce an ensemble of $50$ descriptions per class. 
When generating contrastively, 
we use a temperature of $1.5$ to generate an ensemble of $20$ descriptions per class. 
Like \citet{cupl}, we observe that performance saturates around $50$ descriptions for normal V-GLOSS, 
but we also observe saturation at around $20$ descriptions for contrastive V-GLOSS.
Based on tuning on development data, 
we set $N = 5$, $\lambda = 0.5$, and $k = 4$ (see Algorithm \ref{alg:contrastive}). 
In total, we obtain $70$ class descriptions. 
During generation, we set the maximum number of tokens to $35$, 
but also terminate generation when the \textit{boundary parameter} or \textit{newline} token is reached.

\subsubsection{Results}

The results from Experiment 2, as shown in Table \ref{table:main}, primarily underscore the significant efficiency and accuracy gains of V-GLOSS (\textit{Normal + Contrastive}) over CuPL.

Key findings include:
(1)~V-GLOSS demonstrates an average accuracy improvement of 4.4\% over the baseline (3.3\% for ViT and 5.6\% for RN50).
(2)~Compared to Template Ensembling, V-GLOSS shows an average improvement of 2.2\%.
(3)~Against the variant: V-GLOSS (\textit{Normal}-Only), V-GLOSS (\textit{Normal + Contrastive}) improves accuracy by an average of 1.8\%.

The standout improvements, however, show in the comparison between CuPL and V-GLOSS. Despite having 28.7 times fewer LM parameters than CuPL (6.1B vs. CuPL's 175B), V-GLOSS exhibits notable performance improvements, increasing by an average of 1.8\% on ImageNet, 2.6\% on FGVC Aircraft, 1.6\% on Flowers 102, and 1.4\% across all datasets. The fine-grained datasets (FGVC Aircraft and Flowers 102) show an average improvement of 2.1\%, compared to 0.9\% on the general datasets (ImageNet, CIFAR-10, and CIFAR-100), a nod to the effectiveness of our contrastive algorithm. For a detailed discussion on the implications of our findings, see Section \ref{sec:discussion}.

\subsubsection{Analysis}

In Section~\ref{sec:intro}, we pointed out several problems in previous methods. 
Here, we carefully analyze how V-GLOSS addresses these issues.

\paragraph{Label Ambiguity:} 
Without adequate context, text models may fail to grasp the intended meaning of a polysemous word. 
\textit{Crane} is a polysemous word, and ImageNet \cite{imagenet} has two classes that refer to different senses of the word: \textit{construction machine} and \textit{wading bird}. However, they both use the same label. 
Thus, in \textit{1-Template}, for example, both classes have the same description. 
This point highlights an important benefit of linking classes to WordNet, which resolves such ambiguities.
Empirically, when compared with a ViT backbone to the \textit{Lex Baseline}, our accuracies on \textsc{\textcolor{blue}{crane}} (machine) and \textsc{\textcolor{red}{crane}} (bird) increase from 0\% and 46\% to 76\% and 78\%,  respectively.

\paragraph{Performance-Context Relationship:}
When comparing the baselines to the other methods, we observe that accuracy generally improves as the amount of surrounding context increases. 
On one hand, if a sentence consists of \textit{``my crane.''} alone, the sense of \textit{crane} is unclear. 
On the other, if the sentence is \textit{``my construction crane,''} the meaning of \textit{crane} becomes clearer. 
We see that providing additional context helps to disambiguate words. 
When a description provides more useful context, models can form better representations of specific classes. 
By comparing V-GLOSS to the baselines (see Table \ref{table:main}), we can observe that the benefits of additional context extend to the vision-language setting. 
Concretely, providing visually-grounded context in the description improves performance.

\paragraph{Class Granularity:} 
We consider pairs of classes that are similar enough to be mistaken, such as \textsc{alligator} and \textsc{crocodile}. 
In WordNet, relationships between synsets are modeled through \textit{is-a} (hyponymy-hypernymy) and \textit{part-of} (meronymy-holonymy) relationships. 
For example, \textsc{crocodilian} is a hypernym of both \textsc{alligator} and \textsc{crocodile}, 
while only \textsc{alligator} is a holonym of \textsc{snout}, since alligators have snouts while crocodiles do not.
%
Using our contrastive algorithm, we generate descriptions that highlight how images of a \textsc{crocodile}
should depict a greener animal with a rounded snout. 
Empirically, using ViT, the average accuracy of V-GLOSS across these two classes jumps from 36\% to 68\%
when contrastive glosses are used. 
This improvement highlights the effectiveness of our contrastive V-GLOSS variant 
in reducing false positives between visually similar classes.

\section{Discussion}
\label{sec:discussion}

When looking at our results, a pertinent question arises: 
Why does an SKB, such as WordNet, help us do better on tasks related to vision?
In this section, 
we formulate two insights on how the synergy between SKBs and LMs supports our improvements.
%

\paragraph{Insight \#1: SKBs represent concepts precisely} 
When LMs are prompted with higher-quality context, they produce better output \cite{retro}. 
WordNet provides a precise representation of a class and its relationship to other classes, leaving minimal room for ambiguity. 
Afterward, we can prompt an LM with this precise information to produce unambiguous and high-quality class descriptions.

\paragraph{Insight \#2: Semantic similarity is a useful proxy for visual similarity} 
WordNet models lexical semantics as
a tree (see Figure \ref{fig:contrastive_tree}), with synsets as nodes and \textit{is-a} relationships as directed edges. 
The distance between different nodes reflects the level of semantic similarity, and is by extension an indicator of the level of visual similarity between synsets. 
\textsc{alligator} and \textsc{crocodile} are semantically similar because they are both kinds of \textsc{crocodilian}, but they are visually similar as well (see Table \ref{table:constrastive_descriptions}).
Semantic similarity 
informs what classes we distinguish with our contrastive descriptions, and why they work (see Table \ref{table:class_similarity}).
This is because semantic and visual similarity are highly correlated. 


\section{Conclusion}

This study concentrates on generating visual class descriptions for zero-shot vision tasks. 
We employ a novel method that combines pre-trained language models (LMs) and semantic knowledge bases (SKBs) to create high-quality visual descriptions. 
Our findings suggest that the semantic information from an SKB can condition an LM to generate improved visual descriptions which yield higher accuracy and expressiveness. 
We also show that our contrastive algorithm improves fine-grained discrimination between similar concepts.
The integration of SKBs with LMs reveals partially latent knowledge about visual attributes in the latter and demonstrates a significant interplay between the linguistic and visual domains. 
These results also pave the way for future exploration into leveraging text-only LMs in multi-modal tasks.

\section*{Limitations}

\paragraph{The dataset must be mapped to an SKB.}
As described earlier, mapping the dataset to WordNet, although a one-time step, is not fully automatic.
In future work, we look to fully automate this step, possibly by selecting a synset based on the similarity between sample class images and potential senses of the class label.

\paragraph{We are limited in terms of language, dataset class count, and our SKB's size.}
First, our English-focused stance may prove a limiting factor in our method being applied to ZSIC or ZSCIG tasks based in other languages.
Some classes are strongly related to non-English languages.

Second, our largest evaluation dataset, ImageNet \cite{imagenet}, has 1,000 classes, representing just 0.64\% coverage of WordNet. 
We look forward to evaluating our methods on a larger ImageNet set: ImageNet-21k, which would cover 14.06\% of WordNet.

Third, although our method can be applied to BabelNet \cite{babelnet}, which has over 1.5 billion synsets, we focus on WordNet, which has 155,287.
We look to explore alternative SKBs such as BabelNet, or non-English wordnets, both of which offer the benefit of being multilingual.

\section*{Ethics Statement}

In normal use, we discover no direct ethical issues with our method. 
Note, however, that we may inherit ethical problems from the components used by our method.
Both CLIP \cite{clip-bias} and LMs \cite{lm-bias} have independently been shown to exhibit some level of bias. Also, semantic resources such as WordNet \cite{wordnet98} tend to focus on formalized concepts. This poses a problem if our method's use concerns people on the fringes of society.

We noted earlier that our method is mostly English-focused.
This could be a source of bias if our method is applied in a multilingual context.
We ask that people do not apply our method to real-world problems where multilingual knowledge is required. 
There is also the issue of semantic resources for low-resource languages not being extensive enough \cite{low}.   

\section*{Acknowledgements}
This work was supported by the Natural Sciences and Engineering Research Council of Canada (NSERC) and the Alberta Machine Intelligence Institute (Amii). 
Special thanks to Lili Mou for supplementary compute resources, and to Ning Shi for insightful discussions related to the research.


\bibliography{anthology,custom}

\begin{thebibliography}{37}
\expandafter\ifx\csname natexlab\endcsname\relax\def\natexlab#1{#1}\fi

\bibitem[{Agarwal et~al.(2021)Agarwal, Krueger, Clark, Radford, Kim, and Brundage}]{clip-bias}
Sandhini Agarwal, Gretchen Krueger, Jack Clark, Alec Radford, Jong~Wook Kim, and Miles Brundage. 2021.
\newblock Evaluating clip: towards characterization of broader capabilities and downstream implications.
\newblock \emph{arXiv preprint arXiv:2108.02818}.

\bibitem[{Betker et~al.(2023)Betker, Goh, Jing, Brooks, Wang, Li, Ouyang, Zhuang, Lee, Guo et~al.}]{betker2023improving}
James Betker, Gabriel Goh, Li~Jing, Tim Brooks, Jianfeng Wang, Linjie Li, Long Ouyang, Juntang Zhuang, Joyce Lee, Yufei Guo, et~al. 2023.
\newblock Improving image generation with better captions.
\newblock \emph{Computer Science. https://cdn. openai. com/papers/dall-e-3. pdf}.

\bibitem[{Black et~al.(2022)Black, Biderman, Hallahan, Anthony, Gao, Golding, He, Leahy, McDonell, Phang et~al.}]{gpt-neox}
Sid Black, Stella Biderman, Eric Hallahan, Quentin Anthony, Leo Gao, Laurence Golding, Horace He, Connor Leahy, Kyle McDonell, Jason Phang, et~al. 2022.
\newblock Gpt-neox-20b: An open-source autoregressive language model.
\newblock \emph{arXiv preprint arXiv:2204.06745}.

\bibitem[{Borgeaud et~al.(2022)Borgeaud, Mensch, Hoffmann, Cai, Rutherford, Millican, Van Den~Driessche, Lespiau, Damoc, Clark et~al.}]{retro}
Sebastian Borgeaud, Arthur Mensch, Jordan Hoffmann, Trevor Cai, Eliza Rutherford, Katie Millican, George~Bm Van Den~Driessche, Jean-Baptiste Lespiau, Bogdan Damoc, Aidan Clark, et~al. 2022.
\newblock Improving language models by retrieving from trillions of tokens.
\newblock In \emph{International conference on machine learning}, pages 2206--2240. PMLR.

\bibitem[{Brown et~al.(2020)Brown, Mann, Ryder, Subbiah, Kaplan, Dhariwal, Neelakantan, Shyam, Sastry, Askell et~al.}]{gpt3}
Tom Brown, Benjamin Mann, Nick Ryder, Melanie Subbiah, Jared~D Kaplan, Prafulla Dhariwal, Arvind Neelakantan, Pranav Shyam, Girish Sastry, Amanda Askell, et~al. 2020.
\newblock Language models are few-shot learners.
\newblock \emph{Advances in neural information processing systems}, 33:1877--1901.

\bibitem[{Coates et~al.(2011)Coates, Ng, and Lee}]{stl10}
Adam Coates, Andrew Ng, and Honglak Lee. 2011.
\newblock An analysis of single-layer networks in unsupervised feature learning.
\newblock In \emph{Proceedings of the fourteenth international conference on artificial intelligence and statistics}, pages 215--223. JMLR Workshop and Conference Proceedings.

\bibitem[{Deng et~al.(2009)Deng, Dong, Socher, Li, Li, and Fei-Fei}]{imagenet}
Jia Deng, Wei Dong, Richard Socher, Li-Jia Li, Kai Li, and Li~Fei-Fei. 2009.
\newblock Imagenet: A large-scale hierarchical image database.
\newblock In \emph{2009 IEEE conference on computer vision and pattern recognition}, pages 248--255. Ieee.

\bibitem[{Devlin et~al.(2018)Devlin, Chang, Lee, and Toutanova}]{bert}
Jacob Devlin, Ming-Wei Chang, Kenton Lee, and Kristina Toutanova. 2018.
\newblock Bert: Pre-training of deep bidirectional transformers for language understanding.
\newblock \emph{arXiv preprint arXiv:1810.04805}.

\bibitem[{Dosovitskiy et~al.(2020)Dosovitskiy, Beyer, Kolesnikov, Weissenborn, Zhai, Unterthiner, Dehghani, Minderer, Heigold, Gelly et~al.}]{vit}
Alexey Dosovitskiy, Lucas Beyer, Alexander Kolesnikov, Dirk Weissenborn, Xiaohua Zhai, Thomas Unterthiner, Mostafa Dehghani, Matthias Minderer, Georg Heigold, Sylvain Gelly, et~al. 2020.
\newblock An image is worth 16x16 words: Transformers for image recognition at scale.
\newblock \emph{arXiv preprint arXiv:2010.11929}.

\bibitem[{Hao et~al.(2022)Hao, Chi, Dong, and Wei}]{optimizing}
Yaru Hao, Zewen Chi, Li~Dong, and Furu Wei. 2022.
\newblock Optimizing prompts for text-to-image generation.
\newblock \emph{arXiv preprint arXiv:2212.09611}.

\bibitem[{He et~al.(2016)He, Zhang, Ren, and Sun}]{resnet}
Kaiming He, Xiangyu Zhang, Shaoqing Ren, and Jian Sun. 2016.
\newblock Deep residual learning for image recognition.
\newblock In \emph{Proceedings of the IEEE conference on computer vision and pattern recognition}, pages 770--778.

\bibitem[{Heusel et~al.(2017)Heusel, Ramsauer, Unterthiner, Nessler, and Hochreiter}]{fid}
Martin Heusel, Hubert Ramsauer, Thomas Unterthiner, Bernhard Nessler, and Sepp Hochreiter. 2017.
\newblock Gans trained by a two time-scale update rule converge to a local nash equilibrium.
\newblock \emph{Advances in neural information processing systems}, 30.

\bibitem[{Jia et~al.(2021)Jia, Yang, Xia, Chen, Parekh, Pham, Le, Sung, Li, and Duerig}]{align}
Chao Jia, Yinfei Yang, Ye~Xia, Yi-Ting Chen, Zarana Parekh, Hieu Pham, Quoc Le, Yun-Hsuan Sung, Zhen Li, and Tom Duerig. 2021.
\newblock Scaling up visual and vision-language representation learning with noisy text supervision.
\newblock In \emph{International Conference on Machine Learning}, pages 4904--4916. PMLR.

\bibitem[{Krizhevsky et~al.(2009)Krizhevsky, Hinton et~al.}]{cifar}
Alex Krizhevsky, Geoffrey Hinton, et~al. 2009.
\newblock Learning multiple layers of features from tiny images.

\bibitem[{Liang et~al.(2021)Liang, Wu, Morency, and Salakhutdinov}]{lm-bias}
Paul~Pu Liang, Chiyu Wu, Louis-Philippe Morency, and Ruslan Salakhutdinov. 2021.
\newblock Towards understanding and mitigating social biases in language models.
\newblock In \emph{International Conference on Machine Learning}, pages 6565--6576. PMLR.

\bibitem[{Liang et~al.(2022)Liang, Bommasani, Lee, Tsipras, Soylu, Yasunaga, Zhang, Narayanan, Wu, Kumar et~al.}]{helm}
Percy Liang, Rishi Bommasani, Tony Lee, Dimitris Tsipras, Dilara Soylu, Michihiro Yasunaga, Yian Zhang, Deepak Narayanan, Yuhuai Wu, Ananya Kumar, et~al. 2022.
\newblock Holistic evaluation of language models.
\newblock \emph{arXiv preprint arXiv:2211.09110}.

\bibitem[{Magueresse et~al.(2020)Magueresse, Carles, and Heetderks}]{low}
Alexandre Magueresse, Vincent Carles, and Evan Heetderks. 2020.
\newblock Low-resource languages: A review of past work and future challenges.
\newblock \emph{arXiv preprint arXiv:2006.07264}.

\bibitem[{Maji et~al.(2013)Maji, Rahtu, Kannala, Blaschko, and Vedaldi}]{fgvc_aircraft}
Subhransu Maji, Esa Rahtu, Juho Kannala, Matthew Blaschko, and Andrea Vedaldi. 2013.
\newblock Fine-grained visual classification of aircraft.
\newblock \emph{arXiv preprint arXiv:1306.5151}.

\bibitem[{Menon and Vondrick(2022)}]{menon2022visual}
Sachit Menon and Carl Vondrick. 2022.
\newblock Visual classification via description from large language models.
\newblock \emph{arXiv preprint arXiv:2210.07183}.

\bibitem[{Miller(1998)}]{wordnet98}
George~A Miller. 1998.
\newblock \emph{WordNet: An electronic lexical database}.
\newblock MIT press.

\bibitem[{Mokady et~al.(2021)Mokady, Hertz, and Bermano}]{clipcap}
Ron Mokady, Amir Hertz, and Amit~H Bermano. 2021.
\newblock Clipcap: Clip prefix for image captioning.
\newblock \emph{arXiv preprint arXiv:2111.09734}.

\bibitem[{Navigli and Ponzetto(2012)}]{babelnet}
Roberto Navigli and Simone~Paolo Ponzetto. 2012.
\newblock Babelnet: The automatic construction, evaluation and application of a wide-coverage multilingual semantic network.
\newblock \emph{Artificial intelligence}, 193:217--250.

\bibitem[{Nilsback and Zisserman(2008)}]{flowers}
Maria-Elena Nilsback and Andrew Zisserman. 2008.
\newblock \href {https://api.semanticscholar.org/CorpusID:15193013} {Automated flower classification over a large number of classes}.
\newblock \emph{2008 Sixth Indian Conference on Computer Vision, Graphics \& Image Processing}, pages 722--729.

\bibitem[{Ouyang et~al.(2022)Ouyang, Wu, Jiang, Almeida, Wainwright, Mishkin, Zhang, Agarwal, Slama, Ray et~al.}]{instruct-gpt}
Long Ouyang, Jeff Wu, Xu~Jiang, Diogo Almeida, Carroll~L Wainwright, Pamela Mishkin, Chong Zhang, Sandhini Agarwal, Katarina Slama, Alex Ray, et~al. 2022.
\newblock Training language models to follow instructions with human feedback.
\newblock \emph{arXiv preprint arXiv:2203.02155}.

\bibitem[{Pham et~al.(2021)Pham, Dai, Ghiasi, Kawaguchi, Liu, Yu, Yu, Chen, Luong, Wu et~al.}]{basic}
Hieu Pham, Zihang Dai, Golnaz Ghiasi, Kenji Kawaguchi, Hanxiao Liu, Adams~Wei Yu, Jiahui Yu, Yi-Ting Chen, Minh-Thang Luong, Yonghui Wu, et~al. 2021.
\newblock Combined scaling for open-vocabulary image classification.
\newblock \emph{arXiv preprint arXiv: 2111.10050}.

\bibitem[{Pratt et~al.(2022)Pratt, Liu, and Farhadi}]{cupl}
Sarah Pratt, Rosanne Liu, and Ali Farhadi. 2022.
\newblock What does a platypus look like? generating customized prompts for zero-shot image classification.
\newblock \emph{arXiv preprint arXiv:2209.03320}.

\bibitem[{Radford et~al.(2021)Radford, Kim, Hallacy, Ramesh, Goh, Agarwal, Sastry, Askell, Mishkin, Clark, Krueger, and Sutskever}]{clip}
Alec Radford, Jong~Wook Kim, Chris Hallacy, Aditya Ramesh, Gabriel Goh, Sandhini Agarwal, Girish Sastry, Amanda Askell, Pamela Mishkin, Jack Clark, Gretchen Krueger, and Ilya Sutskever. 2021.
\newblock \href {https://doi.org/10.48550/ARXIV.2103.00020} {Learning transferable visual models from natural language supervision}.

\bibitem[{Radford et~al.(2018)Radford, Narasimhan, Salimans, Sutskever et~al.}]{gpt1}
Alec Radford, Karthik Narasimhan, Tim Salimans, Ilya Sutskever, et~al. 2018.
\newblock Improving language understanding by generative pre-training.

\bibitem[{Radford et~al.(2019)Radford, Wu, Child, Luan, Amodei, Sutskever et~al.}]{gpt2}
Alec Radford, Jeffrey Wu, Rewon Child, David Luan, Dario Amodei, Ilya Sutskever, et~al. 2019.
\newblock Language models are unsupervised multitask learners.
\newblock \emph{OpenAI blog}, 1(8):9.

\bibitem[{Rombach et~al.(2022)Rombach, Blattmann, Lorenz, Esser, and Ommer}]{stable}
Robin Rombach, Andreas Blattmann, Dominik Lorenz, Patrick Esser, and Bj{\"o}rn Ommer. 2022.
\newblock High-resolution image synthesis with latent diffusion models.
\newblock In \emph{Proceedings of the IEEE/CVF Conference on Computer Vision and Pattern Recognition}, pages 10684--10695.

\bibitem[{Salimans et~al.(2016)Salimans, Goodfellow, Zaremba, Cheung, Radford, and Chen}]{inception_s}
Tim Salimans, Ian Goodfellow, Wojciech Zaremba, Vicki Cheung, Alec Radford, and Xi~Chen. 2016.
\newblock Improved techniques for training gans.
\newblock \emph{Advances in neural information processing systems}, 29.

\bibitem[{Song et~al.(2022)Song, Dong, Zhang, Liu, and Wei}]{clipfs}
Haoyu Song, Li~Dong, Wei-Nan Zhang, Ting Liu, and Furu Wei. 2022.
\newblock Clip models are few-shot learners: Empirical studies on vqa and visual entailment.
\newblock \emph{arXiv preprint arXiv:2203.07190}.

\bibitem[{Szegedy et~al.(2015)Szegedy, Liu, Jia, Sermanet, Reed, Anguelov, Erhan, Vanhoucke, and Rabinovich}]{inception}
Christian Szegedy, Wei Liu, Yangqing Jia, Pierre Sermanet, Scott Reed, Dragomir Anguelov, Dumitru Erhan, Vincent Vanhoucke, and Andrew Rabinovich. 2015.
\newblock Going deeper with convolutions.
\newblock In \emph{Proceedings of the IEEE conference on computer vision and pattern recognition}, pages 1--9.

\bibitem[{Wei et~al.(2022)Wei, Tay, Bommasani, Raffel, Zoph, Borgeaud, Yogatama, Bosma, Zhou, Metzler et~al.}]{emergent}
Jason Wei, Yi~Tay, Rishi Bommasani, Colin Raffel, Barret Zoph, Sebastian Borgeaud, Dani Yogatama, Maarten Bosma, Denny Zhou, Donald Metzler, et~al. 2022.
\newblock Emergent abilities of large language models.
\newblock \emph{arXiv preprint arXiv:2206.07682}.

\bibitem[{Wu and Palmer(1994)}]{wu_palmer}
Zhibiao Wu and Martha Palmer. 1994.
\newblock Verb semantics and lexical selection.
\newblock \emph{arXiv preprint cmp-lg/9406033}.

\bibitem[{Yu et~al.(2022)Yu, Wang, Vasudevan, Yeung, Seyedhosseini, and Wu}]{coca}
Jiahui Yu, Zirui Wang, Vijay Vasudevan, Legg Yeung, Mojtaba Seyedhosseini, and Yonghui Wu. 2022.
\newblock Coca: Contrastive captioners are image-text foundation models.
\newblock \emph{arXiv preprint arXiv:2205.01917}.

\bibitem[{Zhou et~al.(2022)Zhou, Yang, Loy, and Liu}]{coop}
Kaiyang Zhou, Jingkang Yang, Chen~Change Loy, and Ziwei Liu. 2022.
\newblock Learning to prompt for vision-language models.
\newblock \emph{International Journal of Computer Vision}, 130(9):2337--2348.

\end{thebibliography}
\bibliographystyle{acl_natbib}

\appendix

\section{Appendices}
The appendices contain Table \ref{tab:imagenet_comparison} which compares WordNet glosses to V-GLOSS descriptions for the first 100 classes in ImageNet.
Next, we show examples of the cases where our most frequent sense heuristic for mapping a class to WordNet failed in Table \ref{tab:classes_senses}.
Finally, in Table \ref{table:main_full}, we show a more detailed variant of Table \ref{table:main} which compares multiple variants of the CLIP backbone. 
The authors of CuPL also combined their method with Template Ensembling.
The resulting method, {CuPL + Template Ensembling},
combines the class descriptions from both methods and leads to marginally better performance.

\subsection{Attention Maps}
\begin{figure}[H]
\centering
\includegraphics[width=0.475\textwidth]{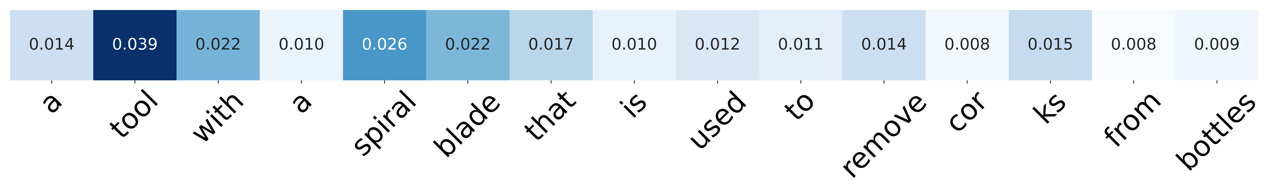}
\caption{Attention map for V-GLOSS description}
\label{fig:vgloss_map}
\end{figure}

\begin{figure}[H]
\centering
\includegraphics[width=0.475\textwidth]{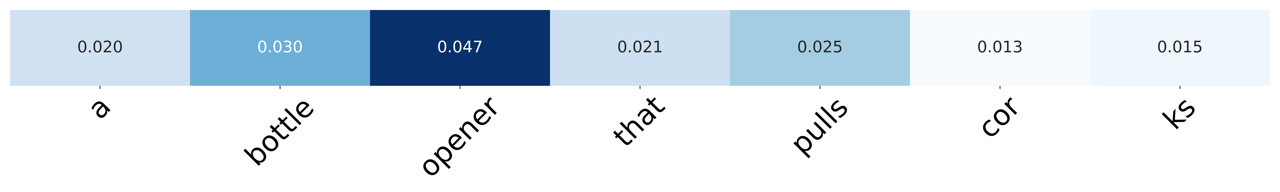}
\caption{Attention map for WordNet gloss}
\label{fig:baseline_map}
\end{figure}

We also briefly analyze V-GLOSS attention maps to better understand its impact on performance. 
Figure \ref{fig:vgloss_map} shows the attention map for V-GLOSS (see Table \ref{table:descript_comparison} for descriptions), indicating effective utilization of visually-relevant context. 
Conversely, Figure \ref{fig:baseline_map} shows the attention map for the WordNet glosses, where the attention score on \textit{bottle} is 3.5x higher, showing less distraction in V-GLOSS. 
These maps demonstrate success in steering the model's attention toward relevant context, thus improving classification accuracy across different classes and descriptions. 
We speculate that our descriptions also reduce distraction in images, but leave this to future work.

\subsection{\emph{Normal} Prompts Per Class}

We also show how accuracy changes with the number of normal prompts used for each class. This result is similar to those shown in \citet{cupl}.
\begin{figure}[H]
\centering
\begin{tikzpicture}
\begin{axis}[
    bar width=.5cm,
    width=\linewidth,
    height=\linewidth,
    legend style={at={(0.5,-0.15)},
    anchor=north,legend columns=-1},
    symbolic x coords={0,1,5,25,50,75,100},
    xtick=data,
    xtick pos=bottom,
    ytick pos=left,
    axis lines=left,
    nodes near coords align={vertical},
    ymin=70,ymax=80,
    ylabel={Accuracy (\%)},
    xlabel={\emph{Normal} Prompts},
    enlarge x limits=0.15
]
\addplot+[mark=*, gray, mark options={fill=black}] coordinates {(1,72.3) (5,74.7) (25,76.2) (50,77.3) (75,77.4) (100,77.4)};
\end{axis}
\end{tikzpicture}
\caption{V-GLOSS Accuracy vs. \emph{Normal} Prompts.}
\label{fig:bar_chart}
\end{figure}
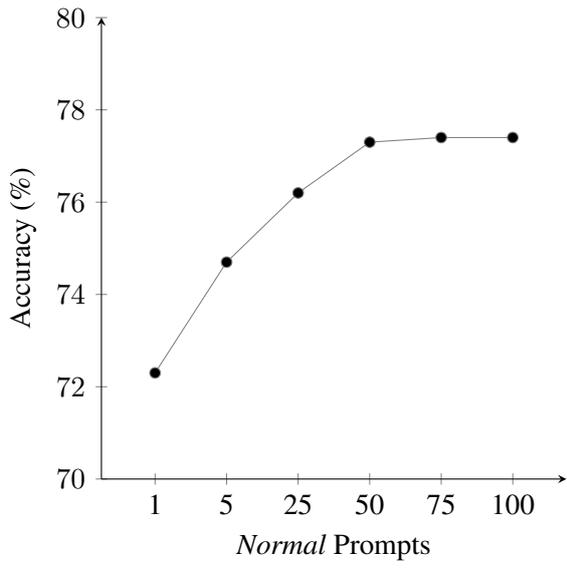

\clearpage
\onecolumn

\subsection{WordNet vs. V-GLOSS}
\captionsetup{
    singlelinecheck = false, 
    width = \textwidth, 
    labelfont = bf, 
    justification = centering
}
{\small
\begin{longtable}{p{2cm} p{6cm} p{6cm}}
\caption{Comparison of WordNet glosses and V-GLOSS descriptions for the first 100 ImageNet classes} 
\label{tab:imagenet_comparison} \\
\toprule
\textbf{Class} & \textbf{WordNet Gloss} & \textbf{V-GLOSS} \\
\midrule
\endfirsthead

\multicolumn{3}{c}%
{{\bfseries Table \thetable\ is continued from previous page}} \\
\midrule
\textbf{Class} & \textbf{WordNet Gloss} & \textbf{V-GLOSS} \\
\midrule
\endhead

\midrule
\multicolumn{3}{r}{{Continued on the next page...}} \\ 
\midrule
\endfoot

\midrule
\endlastfoot

tench & freshwater dace-like game fish of Europe and western Asia noted for ability to survive outside water & Freshwater fish of the family Cyprinidae, found in Europe and western Asia, with a long, slender body, a large mouth, and a forked tail. \\
\hline
goldfish & small golden or orange-red freshwater fishes of Eurasia used as pond or aquarium fishes & Small freshwater fishes of the family Cyprinidae, native to East Asia, introduced to many other parts of the world. \\
\hline
great white shark & large aggressive shark widespread in warm seas; known to attack humans & Large, white, shark-like fish with a large mouth and sharp teeth. \\
\hline
tiger shark & large dangerous warm-water shark with striped or spotted body & Large warm-water shark with a dark brown or black body, white underbelly, and white or yellowish-white dorsal fin. \\
\hline
hammerhead shark & medium-sized live-bearing shark with eyes at either end of a flattened hammer-shaped head; worldwide in warm waters; can be dangerous & Large shark with a flattened head and eyes at either end of the head. \\
\hline
electric ray & any sluggish bottom-dwelling ray of the order Torpediniformes having a rounded body and electric organs on each side of the head capable of emitting strong electric discharges & Sluggish bottom-dwelling rays of the order Torpediniformes with electric organs on each side of the head. \\
\hline
stingray & large venomous ray with large barbed spines near the base of a thin whiplike tail capable of inflicting severe wounds & Large, flat, wingless ray with a long, whip-like tail and sharp, barbed spines near the base. \\
\hline
rooster & adult male bird & Male bird larger than a hen with a crest of feathers on its head. \\
\hline
hen & adult female bird & Bird that lays eggs. \\
\hline
ostrich & fast-running African flightless bird with two-toed feet; largest living bird & Large flightless bird with a long neck and legs, large head, and long, pointed, featherless, yellow-brown or black beak. \\
\hline
brambling & Eurasian finch & Small brown bird with a black head and a white patch on its chest. \\
\hline
goldfinch & small European finch having a crimson face and yellow-and-black wings & Small bird with a bright yellow face and black wings. \\
\hline
house finch & small finch originally of the western United States and Mexico & Small bird with a bright red head, black back, and white breast. \\
\hline
junco & small North American finch seen chiefly in winter & Small brownish-gray bird with a white belly and a dark cap, found in open areas. \\
\hline
indigo bunting & small deep blue North American bunting & Small deep blue North American bunting. \\
\hline
American robin & large American thrush having a rust-red breast and abdomen & Small brown bird with a red breast and a black head and tail. \\
\hline
bulbul & nightingale spoken of in Persian poetry & Small bird with a long tail and a long, pointed beak. \\
\hline
jay & crested largely blue bird & Crested, largely blue bird with a crest on its head and a long tail. \\
\hline
magpie & long-tailed black-and-white crow that utters a raucous chattering call & Black-and-white crow with a long tail, often seen in groups. \\
\hline
chickadee & any of various small grey-and-black songbirds of North America & Small grey-and-black songbird of North America. \\
\hline
American dipper & small stocky diving bird without webbed feet; frequents fast-flowing streams and feeds along the bottom & Small bird with a black head, white breast and back, and white belly; short, thick, black bill and a black tail with white tips. \\
\hline
kite (bird of prey) & any of several small graceful hawks of the family Accipitridae having long pointed wings and feeding on insects and small animals & Large bird with a long pointed tail and a forked tail, used to catch insects and small animals. \\
\hline
bald eagle & a large eagle of North America that has a white head and dark wings and body & Large bird of prey with a white head and dark wings and body. \\
\hline
vulture & any of various large diurnal birds of prey having naked heads and weak claws and feeding chiefly on carrion & Large bird of prey with a bald head, hooked beak, and bare neck. \\
\hline
great grey owl & large dish-faced owl of northern North America and western Eurasia & Large owl with a round head, large eyes, short tail, white face, and gray body with a white patch on the back of the neck. \\
\hline
fire salamander & a kind of European salamander & Small amphibian with a long tail and a long, thin body covered in black and yellow spots. \\
\hline
smooth newt & small semiaquatic salamander & Small semiaquatic salamander with a long tail and a long, pointed snout. \\
\hline
newt & a newt in its terrestrial stage of development & Small amphibian with a long tail, a long, thin body and a short head. \\
\hline
spotted salamander & glossy black North American salamander with yellow spots & Glossy black amphibian with yellow spots. \\
\hline
axolotl & larval salamander of mountain lakes of Mexico that usually lives without metamorphosing & Salamander living in mountain lakes of Mexico, usually found in muddy water. \\
\hline
American bullfrog & largest North American frog; highly aquatic with a deep-pitched voice & Large amphibian with a greenish-brown back and dark brown or black belly; large head with bulging eyes and a long, pointed snout. \\
\hline
tree frog & any of various Old World arboreal frogs distinguished from true frogs by adhesive suckers on the toes & Small frog with a long, thin body, long, thin tail, and long, thin tongue. \\
\hline
tailed frog & western North American frog with a taillike copulatory organ & Small frog with a long, thin tail used for balance and jumping. \\
\hline
loggerhead sea turtle & very large carnivorous sea turtle; wide-ranging in warm open seas & Large, slow-moving, carnivorous sea turtle with a hard shell and long, pointed head. \\
\hline
leatherback sea turtle & wide-ranging marine turtle with flexible leathery carapace; largest living turtle & Large marine turtle with a leathery shell and long, pointed snout. \\
\hline
mud turtle & bottom-dwelling freshwater turtle inhabiting muddy rivers of North America and Central America & Turtle living in muddy rivers and lakes in North America and Central America. \\
\hline
terrapin & any of various edible North American web-footed turtles living in fresh or brackish water & Large, flat-bodied, freshwater turtle with a diamond-shaped shell and long tail. \\
\hline
box turtle & chiefly terrestrial turtle of North America; shell can be closed tightly & Large, slow-moving, terrestrial turtle with a hard shell and long tail. \\
\hline
banded gecko & any of several geckos with dark bands across the body and differing from typical geckos in having movable eyelids; of United States southwest and Florida Gulf Coast & Small lizard with dark bands across its body and a movable eyelid. \\
\hline
green iguana & large herbivorous tropical American arboreal lizards with a spiny crest along the back; used as human food in Central America and South America & Large, bright green lizard with a spiny crest along the back and long tail. \\
\hline
Carolina anole & small arboreal tropical American insectivorous lizards with the ability to change skin color & Small arboreal lizard with a long tail and color-changing skin. \\
\hline
desert grassland whiptail lizard & any of numerous very agile and alert New World lizards & Small lizard with a long tail, usually black and white or brown and white. \\
\hline
agama & small terrestrial lizard of warm regions of the Old World & Small lizards with long tails, long legs, and a long, pointed snout. \\
\hline
frilled-necked lizard & large arboreal insectivorous Australian lizard with a ruff of skin around the neck & Large arboreal insectivorous Australian lizard with a ruff of skin around the neck. \\
\hline
alligator lizard & slim short-limbed lizard having a distinctive fold on each side that permits expansion; of western North America & Slim, short-limbed lizard with a distinctive fold on each side permitting expansion; of western North America. \\
\hline
Gila monster & large orange and black lizard of southwestern United States; not dangerous unless molested & Large, orange and black lizard with a long tail and forked tongue. \\
\hline
European green lizard & a common Eurasian lizard about a foot long & Small reptile with a long tail, pointed snout, and row of spikes along its back. \\
\hline
chameleon & a chameleon found in Africa & Small lizard with a long tail, long neck, and long, thin body covered with many small, sharp scales. \\
\hline
Komodo dragon & the largest lizard in the world (10 feet); found on Indonesian islands & Large lizard with a thick, scaly body, long tail, and large head with sharp teeth. \\
\hline
Nile crocodile & a dangerous crocodile widely distributed in Africa & Large crocodile with a broad, flat snout, long tail, and long, pointed snout. \\
\hline
American alligator & large alligator of the southeastern United States & Large reptile with a long snout, broad head, and long tail. \\
\hline
triceratops & huge ceratopsian dinosaur having three horns and the neck heavily armored with a very solid frill & Large herbivorous dinosaur with three horns and a frill on its neck. \\
\hline
worm snake & small reddish wormlike snake of eastern United States & Small reddish wormlike snake of eastern United States. \\
\hline
ring-necked snake & any of numerous small nonvenomous North American snakes with a yellow or orange ring around the neck & Small nonvenomous snake with a yellow or orange ring around the neck. \\
\hline
eastern hog-nosed snake & harmless North American snake with upturned nose; may spread its head and neck or play dead when disturbed & Harmless North American snake with upturned nose; may spread its head and neck or play dead when disturbed. \\
\hline
smooth green snake & either of two North American chiefly insectivorous snakes that are green in color & Slender, smooth-scaled snake with a green or yellowish-green coloration. \\
\hline
kingsnake & any of numerous nonvenomous North American constrictors; feed on other snakes and small mammals & Large, nonvenomous snake with a pattern of alternating light and dark bands on its body. \\
\hline
garter snake & any of numerous nonvenomous longitudinally-striped viviparous North American and Central American snakes & Slender, smooth-scaled, nonvenomous snake with a long tail and a pattern of alternating light and dark bands. \\
\hline
water snake & any of various mostly harmless snakes that live in or near water & Slender, elongated, usually nonvenomous snake with a flattened head and a long tail often held above the head. \\
\hline
vine snake & slender arboreal snake found from southern Arizona to Bolivia & Slender arboreal snake found from southern Arizona to Bolivia. \\
\hline
night snake & nocturnal prowler of western United States and Mexico & Nocturnal colubrid snake that is black with a white belly and a white stripe down its back. \\
\hline
boa constrictor & very large boa of tropical America and West Indies & Large snake with a thick body and large head covered with scales, forked tongue. \\
\hline
African rock python & very large python of tropical and southern Africa & Large, heavy, non-venomous snake, grey or brown with a black head and yellowish belly. \\
\hline
Indian cobra & a cobra of tropical Africa and Asia & Large venomous snake with a hooded head and forked tongue. \\
\hline
green mamba & green phase of the black mamba & Large venomous snake with a green body and black head and tail. \\
\hline
sea snake & any of numerous venomous aquatic viviparous snakes having a fin-like tail; of warm littoral seas; feed on fish which they immobilize with quick-acting venom & Venomous snake living in the sea with a long, thin body, large head, and forked tongue. \\
\hline
Saharan horned viper & highly venomous viper of northern Africa and southwestern Asia having a horny spine above each eye & Venomous snake with a horny spine above each eye. \\
\hline
eastern diamondback rattlesnake & large deadly rattlesnake with diamond-shaped markings & Large, venomous snake with a diamond-shaped pattern on its back. \\
\hline
sidewinder rattlesnake & small pale-colored desert rattlesnake of southwestern United States; body moves in an s-shaped curve & Small pale-colored desert rattlesnake of southwestern United States; triangular head, long tail, and rattle on its tail. \\
\hline
trilobite & an extinct arthropod that was abundant in Paleozoic times; had an exoskeleton divided into three parts & Fossilized arthropod from the Paleozoic era with a hard exoskeleton divided into three parts. \\
\hline
harvestman & spiderlike arachnid with a small rounded body and very long thin legs & Spiderlike arachnid with a small rounded body and very long thin legs. \\
\hline
scorpion & arachnid of warm dry regions having a long segmented tail ending in a venomous stinger & Venomous arachnid with a segmented tail ending in a stinger. \\
\hline
yellow garden spider & a widely distributed North American garden spider & Large spider with a black body and a yellow abdomen. \\
\hline
barn spider & an orange and tan spider with darkly banded legs that spins an orb web daily & Large, hairy spider with a dark brown body and a white, orange and black pattern on its abdomen. \\
\hline
European garden spider & a spider common in European gardens & Small spider with a long, thin body and a large, round abdomen. \\
\hline
southern black widow & venomous New World spider; the female is black with an hourglass-shaped red mark on the underside of the abdomen & Spider with a black body and a red hourglass-shaped mark on the underside of the abdomen. \\
\hline
tarantula & large hairy tropical spider with fangs that can inflict painful but not highly venomous bites & Large hairy tropical spider with fangs that can inflict painful but not highly venomous bites. \\
\hline
wolf spider & ground spider that hunts its prey instead of using a web & Large, hairy spider with a long, thin body, large head, two large eyes, and a pair of fangs. \\
\hline
tick & any of two families of small parasitic arachnids with barbed proboscis; feed on blood of warm-blooded animals & Small parasitic arachnid that feeds on blood. \\
\hline
centipede & chiefly nocturnal predacious arthropod having a flattened body of 15 to 173 segments each with a pair of legs, the foremost pair being modified as prehensors & Small, segmented, wormlike arthropod with a pair of long, segmented legs and a pair of short, segmented antennae. \\
\hline
black grouse & grouse of which the male is bluish-black & Grouse of which the male is bluish-black. \\
\hline
ptarmigan & large Arctic and subarctic grouse with feathered feet and usually white winter plumage & Large grouse with a white head and neck, brown body, and white tail. \\
\hline
ruffed grouse & valued as a game bird in eastern United States and Canada & Medium-sized game bird with a black body, white breast, and a ruff of feathers around the neck. \\
\hline
prairie grouse & brown mottled North American grouse of western prairies & Large brown mottled North American grouse of western prairies. \\
\hline
peafowl & male peafowl; having a crested head and very large fanlike tail marked with iridescent eyes or spots & Large, colorful, iridescent bird with a fan-shaped tail and a crest on its head. \\
\hline
quail & small gallinaceous game birds & Small game bird with a plump body, short tail, long, pointed bill, and short, rounded tail. \\
\hline
partridge & small Old World gallinaceous game birds & Small bird with a brown body, white breast, and black head and neck. \\
\hline
african grey parrot & commonly domesticated grey parrot with red-and-black tail and white face; native to equatorial Africa & Medium-sized parrots with a grey body, red-and-black tail, and white face. \\
\hline
macaw & long-tailed brilliantly colored parrot of Central America and South America; among the largest and showiest of parrots & Large brightly colored parrot with a long tail and long beak. \\
\hline
sulphur-crested cockatoo & white cockatoo with a yellow erectile crest & Large white cockatoo with a yellow erectile crest. \\
\hline
lorikeet & any of various small lories & Small brightly colored parrot-like bird with a long tail and curved beak. \\
\hline
coucal & Old World ground-living cuckoo having a long dagger-like hind claw & Large bird with a long dagger-like hind claw. \\
\hline
bee eater & colorful chiefly tropical Old World bird having a strong graceful flight; feeds on especially bees & Colorful Old World bird with a strong graceful flight that feeds on bees. \\
\hline
hornbill & bird of tropical Africa and Asia having a very large bill surmounted by a bony protuberance; related to kingfishers & Large tropical bird with a large bill and long tail. \\
\hline
hummingbird & tiny American bird having brilliant iridescent plumage and long slender bills; wings are specialized for vibrating flight & Small bird with a long slender bill and iridescent feathers. \\
\hline
jacamar & tropical American insectivorous bird having a long sharp bill and iridescent green or bronze plumage & Small, colorful birds with long bills and iridescent feathers. \\
\hline
toucan & brilliantly colored arboreal fruit-eating bird of tropical America having a very large thin-walled beak & Large colorful bird with a long beak and crest on its head. \\
\hline
duck & adult male of a wild or domestic duck & Male duck. \\
\hline
red-breasted merganser & widely distributed merganser of America and Europe & Large duck with a red breast and black head and neck. \\
\hline
goose & web-footed long-necked typically gregarious migratory aquatic birds usually larger and less aquatic than ducks & Large bird with a long neck, short tail, usually white with black or brown markings. \\
\hline
\end{longtable}
}

\clearpage
\subsection{Mis-mappings stemming from the Most Frequent Sense Heuristic}
\begin{table*}[h]
\centering
\small
\def\arraystretch{1}
\begin{tabular}{l p{6cm} p{6cm}}
\toprule
\textbf{Class} & \textbf{Wrong Sense} & \textbf{Correct Sense} \\
\midrule
\midrule
Beaver & the soft brown fur of the beaver & large semiaquatic rodent with webbed hind feet and a broad flat tail; construct complex dams and underwater lodges \\
\midrule
Castle & a large and stately mansion & interchanging the positions of the king and a rook \\
\midrule
Cloud & any collection of particles (e.g., smoke or dust) or gases that is visible & a visible mass of water or ice particles suspended at a considerable altitude \\
\midrule
Flatfish & sweet lean whitish flesh of any of numerous thin-bodied fish; usually served as thin fillets & any of several families of fishes having flattened bodies that swim along the sea floor on one side of the body with both eyes on the upper side \\
\midrule
Leopard & the pelt of a leopard & large feline of African and Asian forests usually having a tawny coat with black spots \\
\midrule
Lobster & flesh of a lobster & any of several edible marine crustaceans of the families Homaridae and Nephropsidae and Palinuridae \\
\midrule
Otter & the fur of an otter & freshwater carnivorous mammal having webbed and clawed feet and dark brown fur \\
\midrule
Raccoon & the fur of the North American racoon & an omnivorous nocturnal mammal native to North America and Central America \\
\midrule
Ray & a column of light (as from a beacon) & cartilaginous fishes having horizontally flattened bodies and enlarged winglike pectoral fins with gills on the underside; most swim by moving the pectoral fins \\
\midrule
Seal & fastener consisting of a resinous composition that is plastic when warm; used for sealing documents and parcels and letters & any of numerous marine mammals that come on shore to breed; chiefly of cold regions \\
\midrule
Shrew & a scolding nagging bad-tempered woman & small mouselike mammal with a long snout; related to moles \\
\midrule
Skunk & a person who is deemed to be despicable or contemptible & American musteline mammal typically ejecting an intensely malodorous fluid when startled; in some classifications put in a separate subfamily Mephitinae \\
\midrule
Table & a set of data arranged in rows and columns & a piece of furniture having a smooth flat top that is usually supported by one or more vertical legs \\
\midrule
Television & broadcasting visual images of stationary or moving objects; ;  - Ernie Kovacs & an electronic device that receives television signals and displays them on a screen \\
\midrule
Tiger & a fierce or audacious person & large feline of forests in most of Asia having a tawny coat with black stripes; endangered \\
\midrule
Turtle & a sweater or jersey with a high close-fitting collar & any of various aquatic and land reptiles having a bony shell and flipper-like limbs for swimming \\
\bottomrule
\end{tabular}
\caption{CIFAR-100 classes where the most frequent sense heuristic failed}
\label{tab:classes_senses}
\end{table*}

\clearpage
\subsection{A More Detailed Comparison of Methods Over CLIP variants}
\begin{table*}[h]
\centering
\small
\def\arraystretch{1}
\begin{tabular}{@{}lllllll@{}}
\toprule
\multirow{2.5}{*}{Method} & \multirow{2.5}{*}{Model} & \multicolumn{4}{c}{Datasets} & \multirow{2.5}{*}{\thead{\# LM \\ Parameters}} \\
\cmidrule(l){3-6}
& & {ImageNet} & {CIFAR-100} & {CIFAR-10} & {STL-10} & \\
\midrule
\multirow{2}{*}{Lex Baseline}
         & {ViT-B-32} & 55.7 & 60.5 & 87.4 & 96.3 & \multirow{5}{*}{0} \\
         & {ViT-L-14} & 67.7 & 72.2 & 91.4 & 97.7 & \\
         & {ViT-L-14-336} & 69.1 & 71.9 & 91.5 & 98.2 & \\
         & {RN50} & 51.6 & 34.1 & 69.7 & 91.9 & \\
         & {RN50x64} & 65.9 & 52.6 & 81.1 & 96.4 & \\
\midrule
\multirow{2}{*}{1-Template Baseline}
         & {ViT-B-32} & 59.4 & 64.5 & 88.3 & 97.3 & \multirow{5}{*}{0} \\
         & {ViT-L-14} & 71.1 & 77.3 & 95.2 & 99.5 & \\
         & {ViT-L-14-336} & 72.4 & 76.6 & 94.8 & 99.5 & \\
         & {RN50} & 55.6 & 42.1 & 70.3 & 94.4 & \\
         & {RN50x64} & 68.7 & 57.7 & 81.0 & 98.4 & \\
\midrule
\multirow{2}{*}{Template Ensembling}
         & {ViT-B-32} & 63.2 & 65.1 & 91.3 & 97.2 & \multirow{5}{*}{0} \\
         & {ViT-L-14} & 75.3 & 77.9 & 96.2 & 99.3 & \\
         & {ViT-L-14-336} & 76.2 & 77.5 & 95.7 & 99.4 & \\
         & {RN50} & 59.6 & 41.6 & 75.6 & 94.3 & \\
         & {RN50x64} & 73.2 & 61.3 & 86.8 & 98.3 & \\
\midrule
\multirow{2}{*}{CuPL + Template Ensembling}
         & {ViT-B-32} & 64.6 & - & - & - & \multirow{5}{*}{175B} \\
         & {ViT-L-14} & 76.6 & - & - & - & \\
         & {ViT-L-14-336} & 77.6 & - & - & - & \\
         & {RN50} & 61.3 & - & - & - & \\
         & {RN50x64} & 75.1 & - & - & - & \\
\midrule
\multirow{2}{*}{\citeauthor{menon2022visual}}
         & {ViT-B-32} & 63.0 & - & - & - & \multirow{5}{*}{175B} \\
         & {ViT-L-14} & 75.0 & - & - & - & \\
         & {ViT-L-14-336} & 76.2 & - & - & - & \\
         & {RN50} & - & - & - & - & \\
         & {RN50x64} & - & - & - & - & \\
\midrule
\multirow{2}{*}{V-GLOSS (\textit{Normal}-Only)}
         & {ViT-B-32} & 63.2 & 65.1 & 91.2 & 97.3 & \multirow{5}{*}{6.1B} \\
         & {ViT-L-14} & 75.3 & 76.5 & 95.9 & 99.5 & \\
         & {ViT-L-14-336} & 77.3 & 77.5 & 95.6 & 99.4 & \\
         & {RN50} & 57.9 & 45.6 & 76.7 & 94.3 & \\
         & {RN50x64} & 73.3 & 63.5 & 86.8 & 98.3 & \\
\midrule
\multirow{2}{*}{V-GLOSS (\textit{Normal + Contrastive})}
         & {ViT-B-32} & 65.7 & 66.3 & 92.1 & 97.7 & \multirow{5}{*}{6.1B} \\
         & {ViT-L-14} & 77.6 & \textbf{78.2} & \textbf{97.0} & \textbf{99.6} & \\
         & {ViT-L-14-336} & \textbf{78.5} & 78.0 & 96.0 & \textbf{99.6} & \\
         & {RN50} & 62.8 & 45.8 & 76.8 & 95.0 & \\
         & {RN50x64} & 74.5 & 64.6 & 87.8 & 98.8 & \\
\bottomrule
\end{tabular}
\caption{Top-1 accuracy on ZSIC across five CLIP variants.}
\label{table:main_full}
\end{table*}

\end{document}